\documentclass[journal]{IEEEtran}

\usepackage{xargs}
\usepackage{url}
\usepackage{graphicx}
\usepackage{amsmath} 
\usepackage{amssymb}  
\usepackage{bm}
\usepackage{amsthm}
\usepackage{siunitx}
\usepackage{comment}

\newcommand {\matr}[2]{\left[\hspace{-.5em}\begin{array}{#1}#2\end{array}\hspace{-.5em}\right]}
\newcommand{\x}{{\mathbf{x}}}
\renewcommand{\u}{{\mathbf{u}}}
\newcommand{\vv}{{{v}}}
\newcommand{\w}{{\mathbf{w}}}
\renewcommand{\r}{{\mathbf{r}}}
\newcommand{\rr}{\r}
\newcommand{\W}{\mathcal{W}}
\renewcommand{\figurename}{Figure.}
\newcommand{\ped}{\mathrm{ped}}

\newcommandx{\xb}[2][1=n,2=k]{\x_{#1|#2}}
\newcommandx{\ub}[2][1=n,2=k]{\u_{#1|#2}}
\newcommandx{\vb}[2][1=n,2=k]{\vv_{#1|#2}}
\newcommandx{\wb}[2][1=n,2=k]{\w_{#1|#2}}
\newcommandx{\rx}{\r^{\x}}
\newcommandx{\ru}{\r^{\u}}
\newcommandx{\tb}[2][1=n,2=k]{\tau_{#1|#2}}

\newcommandx{\hb}[3][1=n,2=k,3={}]{h_{#1}^{#3}}
\newcommandx{\gb}[3][1=n,2=k,3={}]{g_{#1|#2}^{#3}}

\newtheorem{Lemma}{Lemma}
\newtheorem{Proposition}{Proposition}

\newtheorem{Assumption}{Assumption}
\newtheorem{Definition}{Definition}
\newtheorem{Remark}{Remark}
\newtheorem{Example}{Example}

\begin{document}
	
	\title{Experimental Validation of Safe MPC for Autonomous Driving in Uncertain Environments}
	\author{Ivo~Batkovic$^{1,2}$, Ankit~Gupta$^{2}$, Mario~Zanon$^3$,~and~Paolo~Falcone$^{1,4}$
		\thanks{This work was partially supported by the Wallenberg Artificial Intelligence, Autonomous Systems and Software Program (WASP) funded by Knut and Alice Wallenberg Foundation.
		}	
		\thanks{$^{1}$ Ivo Batkovic and Paolo Falcone are with the Mechatronics group at the Department of Electrical Engineering, Chalmers University of Technology, Gothenburg, Sweden {\tt\footnotesize \{ivo.batkovic,falcone\}@chalmers.se }}%
		\thanks{$^{2}$ Ivo Batkovic, and Ankit Gupta are with the research department at Zenseact AB {\tt\footnotesize \{ivo.batkovic,ankit.gupta\}@zenseact.com}}%
		\thanks{$^{3}$ Mario Zanon is with the IMT School for Advanced Studies Lucca {\tt\footnotesize mario.zanon@imtlucca.it}}
		\thanks{$^{4}$ Paolo Falcone is with the Dipartimento di Ingegneria ``Enzo Ferrari'' Universit\`a di Modena e Reggio Emilia, Italy {\tt\footnotesize falcone@unimore.it}}}

	\markboth{Journal of \LaTeX\ Class Files,~Vol.~14, No.~8, August~2015}%
	{Shell \MakeLowercase{\textit{et al.}}: Bare Demo of IEEEtran.cls for IEEE Journals}

	\maketitle
	
	\begin{abstract}
		The full deployment of autonomous driving systems on a worldwide scale requires that the self-driving vehicle be operated in a provably safe manner, i.e., the vehicle must be able to avoid collisions in any possible traffic situation. In this paper, we propose a framework based on Model Predictive Control~(MPC) that endows the self-driving vehicle with the necessary safety guarantees. In particular, our framework ensures constraint satisfaction at all times, while tracking the reference trajectory as close as obstacles allow, resulting in a safe and comfortable driving behavior. To discuss the performance and real-time capability of our framework, we provide first an illustrative simulation example, and then we demonstrate the effectiveness of our framework in experiments with a real test vehicle.
	\end{abstract}
	
	
	\IEEEpeerreviewmaketitle
	
	\section{Introduction}
	In the past decade, autonomous systems have attracted the attention of many researchers and companies on a global scale. The research questions and problems posed by the autonomous driving field have in particular been in the research focus, since there is a great potential in enabling such technologies. Already today, Advanced Driver Assist Systems~(ADAS) are being available for highway-like situations, and have been proven to reduce the frequency and severity of some traffic accidents. To that end, it is believed that fully autonomous driving systems are a key component in reaching the vision of eliminating all fatal road traffic accidents~\cite{lubbe2018predicted,ziegler2014making}.
	
	However, in order to fully deploy highly automated driving technologies in more general settings, such as urban driving, the autonomous system needs not only to reliably sense the surrounding environment, but also to \emph{safely} interact with it. To that end, the open problems related to safe autonomous driving span from the design of cost-effective, robust, and reliable sensors (e.g., cameras, LIDARs, radars, GPS, HD-maps) to the development of robust perception and motion planning and control algorithms. In this paper, we assume that the state-of-the-art sensing and localization algorithms are solved, and only consider the open problem of ensuring \emph{safe} autonomous driving in uncertain environments where the future motion of other road users cannot be known a-priori, but must instead be estimated.
	
	To solve the problem of ``how'' self-driving vehicles should behave, i.e., the planning and control of the vehicle motion, optimization-based techniques have proven to be a favorable choice. In particular, Model Predictive Control~(MPC) has been commonly used since it accommodates the control of (non)linear systems that can be subject to time-varying references and (non)linear constraints. Common approaches in the literature use MPC to approach the problems from different point of views, such as: planning and control~\cite{batkovic2020robust,andersson2018receding,batkovic2019real,cesari2017scenario,chen2021interactive,gros2020linear,lima2018experimental,nair2021stochastic,ugo,ljungqvist2018stability,oliveira2019path}, energy consumption minimization~\cite{Hult2018a,uebel2019two,zanon2020gauss,murgovski2018}, and optimal coordination~\cite{de2017traffic,campos2014cooperative,hult2018optimal}. In most of these settings, a reference trajectory (or path) is already given, and the optimization problem is designed to minimize a performance criterion while satisfying constraints that are either imposed by the internal system or by the environment. Notable examples include: (a) maneuvering large vehicles in highly constrained environments~\cite{oliveira2019path,evestedt2016motion,7995817}, (b) controlling vehicles in an intersection safely and efficiently~\cite{Hult2018a,hult2018optimal}, and (c) controlling a vehicle in environments with uncontrollable moving obstacles~\cite{andersson2018receding,batkovic2019real}. However, while it is well-known how to design an MPC controller such that closed-loop stability w.r.t.~a reference trajectory (or path) is obtained and that constraint satisfaction holds at all times~\cite{borrelli2017predictive,rawlings2009model}, the results in the literature rely on assumptions that can be challenging, or even impossible, to satisfy for practical autonomous driving settings. In other words, one cannot directly apply the developed MPC theory to ensure that a self-driving vehicle will make safe decisions at all times, i.e., ensuring \emph{recursive feasibility} of the MPC controller still remains an open problem for environments where other non-controllable road users are present. We note that while the work in~\cite{hult2018optimal} showed that it is possible to formulate conditions for safe optimal coordination of multiple vehicles in intersections, it relied on an assumption that there exists a ``coordinator'' that decides how and when each vehicle should enter the intersection. To that end, such an approach is not directly applicable, or practical, for general autonomous driving settings, since a self-driving vehicle needs to interact with other human-driven vehicles, cyclists, and pedestrians. 
	
	In order to also account for unforeseen events, Contingency Model Predictive Control~(CMPC) was introduced in~\cite{alsterda2019contingency,alsterda2021contingency}, where the MPC problem instead predicts two trajectories that are connected by their first input, where one of the trajectories acts as a contingency plan accounting for a worst-case outcome, while the other trajectory considers the nominal situation. The authors have shown both through simulations and real experiments that the CMPC framework can serve as a credible strategy to augment a deterministic MPC controller with robustness. However, while the additional contingency trajectory helps against unforeseen events, the CMPC framework does not provide recursive feasibility guarantees for the autonomous driving setting. In a similar light,~\cite{brudigam2021stochastic,brudigam2022safe} formulate a two-step approach that accounts for safety. In this framework, a Stochastic Model Predictive Control~(SMPC) problem is first solved for the nominal case. Then, the input of the SMPC problem is used only if a backup MPC formulation, which is more conservative and accounts for worst-case outcomes, can verify that it is safe. Otherwise, the trajectory of the backup MPC takes over and steers the system to a safe configuration. While~\cite{alsterda2019contingency,alsterda2021contingency} and~\cite{brudigam2021stochastic,brudigam2022safe} address safety concerns using an MPC framework, they either rely on extending the state space of the MPC problem, or solving two different problems separately.
	
	In this paper, we address the problem of ensuring the safe operation of an autonomous vehicle in uncertain environments, e.g., how a self-driving vehicle should be operated so that it never collides with other road users. We base our approach on the novel safe Model Predictive Flexible trajectory Tracking Control~(MPFTC) framework~\cite{batkovic2020safe} and show in practice how a safe vehicle controller can be designed for urban autonomous driving settings. In particular, our contributions are the following. First, we introduce an MPC problem formulation that ensures robust constraint satisfaction for the self-driving vehicle, while only slightly modifying the standard MPC design procedure and relying on a set of (verifiable) assumptions. Note that, contrary to our contribution,~\cite{alsterda2019contingency,alsterda2021contingency,brudigam2021stochastic,brudigam2022safe} do not provide a general framework for safe autonomous driving. In addition, we show through simulations that safety, even in trivial traffic situations, can become jeopardized with an MPC controller if the required assumptions do not satisfy the safe MPC formulation. Finally, by deploying the safe MPC framework on a real test vehicle, we show through a real experiment the performance and real-time capabilities of the framework.
	
	\textit{Contributions of the paper:} The work presented in~\cite{batkovic2020safe} provides a generic safe framework which is not trivial to apply to the case of autonomous driving. This paper shows how to specialize the generic formulation in~\cite{batkovic2020safe} to urban autonomous driving situations, and that it can indeed be used to guarantee safety. In particular, we show the effectiveness of the framework in simulations, and also deploy it on a real vehicle.
	
	\textit{Paper structure:} Section~\ref{sec:problem} introduces the MPC problem statement and the assumptions needed to satisfy safety of the control algorithm, while Section~\ref{sec:implementation} discusses and shows how to enforce all required assumptions in practice. In particular, Section~\ref{sec:system} introduces the considered system model and controller details, Section~\ref{sec:collision_avoidance} discusses how to model the a-priori unknown constraints in order to ensure robust constraint satisfaction, and Section~\ref{sec:terminal} shows how to derive the terminal conditions needed to ensure recursive feasibility of the controller. In Sections~\ref{sec:simulation} and~\ref{sec:experiments} we show the performance of the proposed controller through simulations and real experiments, respectively. Finally, in Section~\ref{sec:conclusions} we draw conclusions and outline future work directions.

	\section{Problem Formulation}\label{sec:problem}
	Our objective is to design a controller for urban autonomous driving applications that allows for comfortable driving in environments where other road users are present, e.g., pedestrians, cyclists, and other vehicles. Before we state the problem formulation, we first introduce terms and notations that will be used throughout the paper.
	
	\subsection{Notation}
	
	We consider a discrete-time nonlinear system defined by
	\begin{equation}\label{eq:nonlinear_model}
		\x_{k+1} = f(\x_k,\u_k),
	\end{equation}
	where $\x_k\in\mathbb{R}^{n_\x}$ denotes the state and $\u_k\in\mathbb{R}^{n_\u}$ denotes the control input at time $k$. The system is subject to state and input constraints of two categories: a-priori known constraints $h(\x,\u):\mathbb{R}^{n_\x}\times\mathbb{R}^{n_\u}\rightarrow\mathbb{R}^{n_h}$, which might be time-varying but which are fully known beforehand; and a-priori unknown constraints $g(\x,\u):\mathbb{R}^{n_\x}\times\mathbb{R}^{n_\u}\rightarrow\mathbb{R}^{n_g}$, whose functional form is known a priori, but whose value is not. In both cases, the state and input must satisfy $h(\x,\u)\leq{}0$, $g(\x,\u)\leq{}0$ and all inequalities are defined element-wise. Examples of a-priori known constraints include engine torque limitations and braking force, while examples of a-priori unknown constraints include collision avoidance with other road users.
	
	Due to the nature of the known and unknown constraints, we use $g_{n|k}(\x,\u)$ to denote function $g$ at time $n$, given the information available at time $k$. Moreover, we will denote by $g_n(\x,\u):=g_{n|\infty}(\x,\u)=g_{n|k}(\x,\u)$, $\forall \, k \geq n$ the real constraint, since in general $\gb[n][k](\x,\u)\neq g_n(\x,\u)$, $\forall\, k<n$. Note that instead for a-priori known constraints $h_{n|k}(\x,\u):=h_n(\x,\u)$ holds $\forall \, k$ by definition. Throughout the remainder of the paper, we also apply the same notation to the predicted state and inputs, e.g., $\xb$ and $\ub$ denote the predicted state and input at time $n$ given the current time $k$. In addition, we will use $\mathbb{I}_a^b:=\{a,a+1,...,b\}$ to denote a set of integers.
	
	Finally, we provide the following definition of safety which will be the foundation of our safety argumentation.
	\begin{Definition}[Safety]\label{def:safe}
		A controller is said to be safe in a given set $\mathcal{S}\subseteq\mathbb{R}^{n_\x}$ if $\forall \, \x_0\in\mathcal{S}$ if it is robust control invariant for the closed-loop system, i.e., it generates control inputs $\mathbf{U}=\{\u_0,...,\u_\infty\}$ and corresponding state trajectories $\mathbf{X}=\{\x_0,\x_1,...,\x_\infty\}$ such that $h_k(\x_k,\u_k)\leq{}0$ and~$g_k(\x_k,\u_k)\leq{}0$, $\forall \, k \geq 0$.
	\end{Definition}
	
	\subsection{Problem Statement}
	Since our objective is to design a controller that is safe and comfortable, the controller must: (a) guarantee safety, in the sense of Definition~\ref{def:safe}, at all times; and (b) control system \eqref{eq:nonlinear_model} such that the state and input $(\x_k,\u_k)$ track a user-defined parameterized reference $\r(\tau):=(\rx(\tau),\ru(\tau))$ as closely as possible. Parameter $\tau$ can be interpreted as a ``fictitious time'', as a natural choice for the reference in MPC is to use time as a parameter. Note that, if $\tau$ is selected to be time, its natural dynamics are given by
	\begin{equation}
		\tau_{k+1}= \tau_{k} + t_\mathrm{s},
	\end{equation}
	where $t_\mathrm{s}$ is the sampling time for sampled-data systems and $t_\mathrm{s}=1$ in the discrete-time framework.
	
	To ensure that requirements (a) and (b) are satisfied in the urban autonomous driving setting, we use the framework proposed in~\cite{batkovic2020safe}, where the idea is to adjust the reference trajectory in order to avoid aggressive tracking behaviors. This is done by adapting the dynamics of the reference trajectory by the means of the parameter $\tau$, which acts as a fictitious time through the dynamics given by
	\begin{equation}
		\tau_{k+1} = \tau_k + t_\mathrm{s} + v_k,
	\end{equation}
	where $v$ is an additional auxiliary control input, and $\tau$ comes as an auxiliary state. With this relaxed timing law of the reference trajectory, we proceed to formulate the practical version of Model Predictive Flexible trajectory Tracking Control~(MPFTC)~\cite[Section V.A]{batkovic2020safe} through the following problem formulation
	\begin{subequations}\label{eq:nmpc}
		\begin{align}
			\min_{\substack{\x\\\tau},\substack{\u\\v}}& \sum_{n=k}^{k+N-1}
			q_\r(\xb,\ub,\tb)+w \vb^2&\\ 
			& &\hspace{-10em}+p_\r(\xb[k+N],\tb[k+N])\nonumber\\
			\text{s.t.}\ &\xb[k][k] = \x_{k},\  \tb[k] = \tau_{k},&\label{eq:nmpc_initial} \\
			&\xb[n+1] = f(\xb,\ub), &\hspace{-1em}n\in \mathbb{I}_k^{k+M-1},\label{eq:nmpc_fx}\\
			&\tb[n+1] = \tb+t_\mathrm{s}+\vb, &\hspace{-1em}n\in \mathbb{I}_k^{k+M-1},\label{eq:nmpc_ft}\\
			&\hb(\xb,\ub) \leq{} 0, & \hspace{-1em}n\in \mathbb{I}_k^{k+M-1},\label{eq:nmpc_h}\\
			&\gb[n][k](\xb,\ub) \leq{} 0, & \hspace{-1em}n\in \mathbb{I}_k^{k+M-1},\label{eq:nmpc_g}\\
			&\xb[k+n] \in\mathcal{X}_\r^\mathrm{s}(\tb[k+n]),&\hspace{-1em}n\in \mathbb{I}_{k+N}^{k+M-1},\label{eq:nmpc_terminal}\\
			&\xb[k+M]\in\mathcal{X}_\mathrm{safe}(\tb[k+M])\subseteq\mathcal{X}_\r^\mathrm{s}(\tb[k+M]).\hspace{-10em}&\label{eq:nmpc_safe_terminal}
		\end{align}
	\end{subequations}
	where $k$ is the current time, $N$ is the prediction horizon associated with a cost, $M\geq{}N$ is the full prediction horizon, $q_\r$ and $p_\r$ are cost functions that penalize deviations from the reference trajectory $\r(\tau)=(\r^\x(\tau), \r^\u(\tau))$, and $w>0$ is associated with the cost of the auxiliary input $\vb$. The predicted state and control inputs are denoted by $\xb$ and $\ub$, respectively. Constraint~\eqref{eq:nmpc_initial} ensures that the prediction starts at the current state, while constraints~\eqref{eq:nmpc_fx}-\eqref{eq:nmpc_ft} enforce the system dynamics. Constraints~\eqref{eq:nmpc_h} represent known constraints, e.g., state and actuation limits, while constraints~\eqref{eq:nmpc_g}  represent collision-avoidance w.r.t other road users, e.g., pedestrians, cyclists and other vehicles. Finally, constraints~\eqref{eq:nmpc_terminal} and~\eqref{eq:nmpc_safe_terminal} define a stabilizing set and a safe set, respectively. In order to provide additional intuition behind this problem formulation, we point out next the main differences between Problem~\eqref{eq:nmpc} and a standard MPC formulation. 
	
	The first difference  is that~\eqref{eq:nmpc} uses the MPFTC framework, i.e., it depends on the auxiliary inputs $\tau$ and $v$, and the terminal sets are parameterized with the parameter $\tau$. The motivation behind this design choice is that adapting $\tau$ offers a flexibility in the tracking behavior whenever the constraints are causing the unmodified reference to become infeasible or the state $\x$ is far from the reference, i.e., it allows the system to behave less aggressive when trying to keep up with the reference trajectory. The artificial reference only changes the cost of the MPC problem, such that its introduction does not change how restrictive the approach is, as the constraints are unaffected. Note that, enforcing the constraint $v=0$ forces the reference trajectory to move as in standard MPC.
	
	The second difference is that Problem~\eqref{eq:nmpc} can be seen as a two stage-problem, where the first stage $n\in\mathbb{I}_{k}^{k+N}$ defines the MPC problem, and the second stage  $n\in\mathbb{I}_{k+N+1}^{k+M}$ defines the terminal set implicitly, where the properties of sets $\mathcal{X}_\r^\mathrm{s}(t)$ and $\mathcal{X}_\mathrm{safe}(t)$ will be stated in Assumptions~\ref{a:stab} and~\ref{a:safe}. This implicit terminal set formulation relates to the safety design which ensures that the state $\x_{k+N|k}$ is able to reach the safe set $\mathcal{X}_\mathrm{safe}(t_{k+M|k})$ in a finite amount of time $M-N\geq{}0$ while satisfying the system dynamics and the a-priori known and unknown constraint. For a more detailed discussion related to the design of Problem~\eqref{eq:nmpc}, we refer the reader to~\cite{batkovic2020safe}.
	
\begin{figure}[t]
		\centering
		\includegraphics[width=\linewidth]{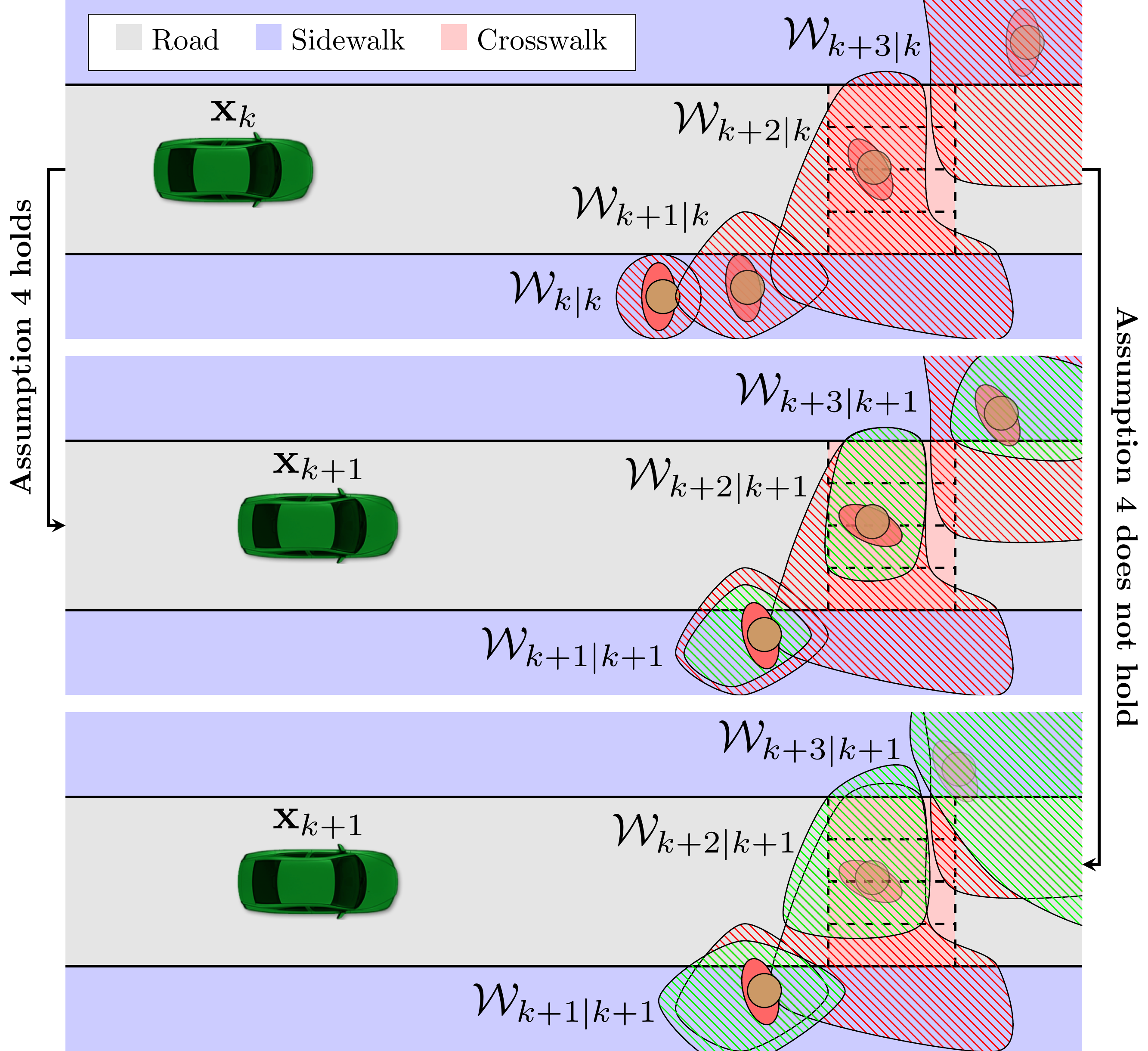}
		\caption{The top panel shows an initial prediction for a pedestrian at time $k$, while the middle and bottom panels show different predictions at time $k+1$. The middle panel illustrates a model that satisfies~\eqref{eq:reachable}, i.e., satisfying Assumption~\ref{a:unknown}, while the bottom panel does not satisfy~\eqref{eq:reachable}, hence, Assumption~\ref{a:unknown} is also not satisfied.}
		\label{fig:consistent}
	\end{figure}

	In order to prove that the controller based on Problem~\eqref{eq:nmpc} is safe, we first need to introduce the following assumptions
	\begin{Assumption}[Regularity]\label{a:cont}
		The system model $f$ is continuous and the stage cost $q_\r:\mathbb{R}^{n_x}\times\mathbb{R}^{n_u}\times\mathbb{R}\rightarrow\mathbb{R}_{\geq{}0}$ and terminal cost $p_\r:\mathbb{R}^{n_x}\times\mathbb{R}\rightarrow\mathbb{R}_{\geq{}0}$, are continuous at reference, and satisfy $q_\r(\rx(t_k),\ru(t_k),t_k)=0$ and $p_\r(\rx(t_k),t_k)=0$. Additionally, $q_\r(\x_k,\u_k,t_k)\geq{}\alpha_1(\|\x_k-\rx(t_k)\|)$ for all feasible $\x_k$, $\u_k$, and $p_\r(\x_k,t_k)\leq{}\alpha_2(\|\x_k-\rx(t_k)\|)$, where $\alpha_1$ and $\alpha_2$ are $\mathcal{K}_\infty$-functions.
	\end{Assumption}
	\begin{Assumption}[Reference feasibility]\label{a:ref}
		The reference trajectory is feasible for the system dynamics, i.e., $\r^\x(t_{k}+t_\mathrm{s})=f(\r^\x(t_k),\r^\u(t_k))$, and the reference satisfies the known constraints~\eqref{eq:nmpc_h}, i.e., $h_n(\r^\x(t_n),\r^\u(t_n))\leq{}0$, for all $n\in\mathbb{I}_0^\infty$.
	\end{Assumption}
	\begin{Assumption}[Stabilizing Terminal Conditions]\label{a:stab} There exists a stabilizing terminal set $\mathcal{X}^\mathrm{s}_\r$ and a terminal control law $\kappa_\r^\mathrm{s}(\x)$ yielding $\x_+^\kappa = f(\x,\kappa_\r^\mathrm{s}(\x))$ and $t_+=t_k+t_\mathrm{s}$, such that $p(\x_+^\kappa,t_+)-p(\x,t)\leq{}-q(\x,\kappa_\r^\mathrm{s}(\x),t)$ and $\x\in\mathcal{X}_\r^\mathrm{s}(t)\Rightarrow \x_+^\kappa\in\mathcal{X}_\r^\mathrm{s}(t_+)$, and $h_n(\x,\kappa_\r^\mathrm{s}(\x))\leq{}0$, for all $n,k\in\mathbb{I}_{0}^\infty$.
	\end{Assumption}
	While Assumptions~\ref{a:cont}-\ref{a:stab} coincide with those commonly used in standard MPC to prove stability, see e.g.,~\cite{rawlings2009model,Grune2011}, we must also introduce the following assumptions in order to gurantee safety for the setting of autonomous driving situations where other road users are present.
	\begin{Assumption}[Unknown constraint dynamics]\label{a:unknown}The a-priori unknown constraint functions satisfy $g_{n|k+1}(\xb,\ub)\leq{}g_{n|k}(\xb,\ub)$ for all $n\geq{}k$.
	\end{Assumption}
	This assumption only requires that the a-priori unknown constraint $g_{n|k}$, which is associated to the predicted uncertainty of other road users, has a consistent characterization. In the context of autonomous driving, function $g$ characterizes, among other constraints, the positions in which no collision with other road users occurs. A position is safe at a given time $n$ if no other road user can be in that position at time $n$. By using a prediction model for other road users, one must consider as unsafe those positions which can be reached by other road users. Assumption~\ref{a:unknown} then requires that the reachable sets predicted at time $k$ contain the corresponding reachable sets predicted at time $k+1$, as shown in Figure~\ref{fig:consistent}. This construction is discussed in depth in Section~\ref{sec:collision_avoidance}. 
	
	The final following assumption is needed to ensure that the controller based on Problem~\eqref{eq:nmpc} is recursively feasible.
	\begin{Assumption}\label{a:safe}
		There exists a robust invariant set denoted $\mathcal{X}_\mathrm{safe}\subseteq\mathbb{R}^{n_x}$ such that for all $\x\in \mathcal{X}_\mathrm{safe}$ there exists a safe control set $\mathcal{U}_\mathrm{safe}\subseteq\mathbb{R}^{n_u}$ entailing that  $f(\x,\u_\mathrm{safe})\in\mathcal{X}_\mathrm{safe}$, and $h_n(\x,\u_\mathrm{safe})\leq{}0$, for all $\u_\mathrm{safe}\in\mathcal{U}_\mathrm{safe}$. Moreover, for all $\x\in \mathcal{X}_\mathrm{safe}$ the a-priori unknown constraints can never be positive, i.e., $\gb[n][k](\x,\u_\mathrm{safe}) \leq 0$ holds for all $\x\in \mathcal{X}_\mathrm{safe}$,  $\u_\mathrm{safe}\in\mathcal{U}_\mathrm{safe}$.
	\end{Assumption}
	Note that Assumption~\ref{a:safe} is the translation of Definition~\ref{def:safe} in mathematical terms, which states that any set of states which are safe must be forward invariant and it must be such that the constraints $\gb$ are not violated. On the contrary, if no such safe configuration exists, then system~\eqref{eq:nonlinear_model} is intrinsically unsafe. Hence, Assumption~\ref{a:safe} is essential for recursive feasibility of the controller based on Problem~\eqref{eq:nmpc}. In the context of autonomous driving, a safe set could involve stopping the car in a suitable location. We will provide more details regarding the safe set in Section~\ref{sec:terminal}.
	
	We are now ready to state the following result.
	\begin{Proposition}[Recursive Feasibility]\label{prop:recursive}
		Suppose that Assumptions~\ref{a:cont}, \ref{a:ref}, \ref{a:stab}, \ref{a:unknown}, and \ref{a:safe} hold, and that Problem~\eqref{eq:nmpc} is feasible for the initial state $(\x_k,t_k)$. Then, system~\eqref{eq:nonlinear_model} in closed loop with the solution of~\eqref{eq:nmpc} applied in receding horizon is safe (recursively feasible) at all times.
		\begin{proof}
			The full proof is given in~\cite[Theorem 2]{batkovic2020safe}.
		\end{proof}
	\end{Proposition}
	
	While this proposition states under which assumptions recursive feasibility (safety) can be ensured, it does not discuss how to enforce them in practice for our considered autonomous driving setting. In the next section we therefore show how to apply the theory from Proposition~\ref{prop:recursive} in practice.
	
	\section{Implementation of Safe MPC}\label{sec:implementation}
	
	In this section we discuss how our proposed framework can be applied in practice and how satisfaction of Assumptions~\ref{a:cont}-\ref{a:safe} can be enforced. In Section~\ref{sec:system} we introduce the vehicle model, cost, and system constraints that satisfy Assumptions~\ref{a:cont} and~\ref{a:ref}. Then, in Section~\ref{sec:collision_avoidance} we show that Assumption~\ref{a:unknown} can be satisfied by using prediction models of the environment to construct the a-priori unknown constraints $\gb$. Finally, in Section~\ref{sec:terminal} we derive the terminal conditions that satisfy Assumptions~\ref{a:stab} and~\ref{a:safe}.
	
	\subsection{Vehicle Model and System Constraints}\label{sec:system}
	We model the vehicle dynamics using the single-track vehicle model
	\begin{equation}\label{eq:vehicle_model}
		\matr{c}{
			\dot{x}\\
			\dot{y}\\
			\dot{\psi}\\
			\dot{\delta}\\
			\dot{\alpha}\\
			\dot{v}\\
			\dot{a}} = 
		\matr{c}{
			v\cos(\psi)\\
			v\sin(\psi)\\
			\frac{v}{l}\tan(\delta)\\
			\alpha\\
			w_0^2(\delta^{\mathrm{sp}}-\delta) - 2w_0w_1\alpha\\
			a\\
			t_\mathrm{acc}(a^{\mathrm{req}}-a)},
	\end{equation}
	where $x,y$ denote the position coordinates in a global frame, $\psi$ is the orientation angle, $\delta$ is the steering angle, $\alpha$ is the steering angle rate, $v$ is the velocity, and $a$ is the acceleration. The control inputs are given by the acceleration request $a^\mathrm{req}$ and steering angle setpoint $\delta^\mathrm{sp}$. Finally, $w_0$, $w_1$ and $t_\mathrm{acc}$ are model constants defining the steering actuator and acceleration dynamics.
	
	In order to track the user-defined reference $\r$, we express the vehicle kinematics in the frame of the reference path~\cite{lima2017stability}
	\begin{align}\label{eq:error_model}
		\matr{c}{
			\dot{s}\\
			\dot{e}_y\\
			\dot{e}_\psi\\
			\dot\delta\\
			\dot\alpha\\
			\dot{v}\\
			\dot{a}
		} 
		&=\left[
		\begin{array}{ c }
			v\cos(e_\psi)(1-\kappa^\rr(s)e_y)^{-1}\\ \hline
			v\sin(e_\psi)\\
			v l^{-1}\tan(\delta)-\dot{s}l^{-1}\tan(\delta^\rr(s))\\
			\alpha\\
			w_0^2(\delta^{\mathrm{sp}}-\delta) - 2w_0w_1\alpha\\ \hline
			a\\
			t_\mathrm{acc}(a^{\mathrm{req}}-a)
		\end{array}
		\right] =
		\left[
		\begin{array}{ c }
			\text{aux.}\\ \hline
			\\
			\text{lat.}\\
			\text{kin.}\\
			\\ \hline
			\text{lon.}\\
			\text{kin.}\\
		\end{array}
		\right], \nonumber
		\\
		\x&=\matr{cccccc}{
			e_y&
			e_\psi&
			\delta&
			\alpha& v&
			a
		}^\top,\ 
		\u=
		\matr{cc}{
			a^\mathrm{req}&
			\delta^\mathrm{sp}
		}^\top
	\end{align}
	where $s$ is the longitudinal position along the path, $\kappa^\mathrm{r}(s)$ is the path curvature along the path, $e_y$ is the lateral displacement error, $e_\psi$ is the yaw error with respect to the reference $\r$ and $\delta^\rr$ is the reference steering angle. Since the reference in~\eqref{eq:error_model} becomes parameterized along the longitudinal position $s$, we consider $s$ to be an auxiliary state (as opposed to $\tau$ in~\eqref{eq:nmpc}) which we do not track, as we are only interested in tracking the velocity while remaining on the reference.
	
	For both simulations and experiments in Sections~\ref{sec:simulation} and~\ref{sec:experiments}, we consider that system~\eqref{eq:error_model} is subject to the following known constraints
	\begin{gather*}
		\| e_y \| \leq{} \bar{e}_y,\ \| e_\psi\| \leq{} \bar{e}_\psi,\  \| \delta \| \leq{} \bar{\delta},\ \| \delta^\mathrm{sp} \| \leq{} \bar{\delta},\\
		0\leq{}v\leq{}\bar{v},\  \underline{a}\leq{}a\leq{}\bar{a},\ -\underline{a}\leq{}a^\mathrm{req}\leq{}\bar{a},\ \| \alpha\|\leq{}\bar{\alpha}.
	\end{gather*}
	The costs $q_\r$ and $p_\r$ are chosen as quadratic costs with their minimum on the reference $\r=(\rx,\ru)$, i.e,
	\begin{gather*}
		q_\r(\x_k,\u_k,s_k) := \matr{c}{\x_k-\rx(s_k)\\\u_k-\ru(s_k)}^\top{}W\matr{c}{\x_k-\rx(s_k)\\\u_k-\ru(s_k)},\\
		p_\r(\x_k,s_k) := (\x_k-\rx(s_k))^\top{}P(\x_k-\rx(s_k)),
	\end{gather*}
	where $W:=\mathrm{blockdiag}(Q,R)$ is constructed using positive-definite matrices $Q\in\mathbb{R}^{6\times6}$ and $R\in\mathbb{R}^{2\times2}$. The computation of the terminal cost matrix $P\in\mathbb{R}^{6\times6}$ is discussed in the Section~\ref{sec:terminal}.
	
	The next step in formulating Problem~\eqref{eq:nmpc} is to construct the a-priori unknown constraints $\gb$. Therefore, we provide next a discussion on how prediction models of the surrounding environment can be used.

	\subsection{Predictive Collision Avoidance Constraints}\label{sec:collision_avoidance}
	In order to apply the results from Proposition~\ref{prop:recursive}, one must first understand how constraints $g_{n|k}$ can be modeled such that Assumption~\ref{a:unknown} is satisfied. This section will therefore provide a formal description of how $g_{n|k}$ can be constructed, and then show how a pedestrian prediction model can be derived.

	We introduce function $\gamma(\x,\u,\w):\mathbb{R}^{n_\x}\times\mathbb{R}^{n_\u}\times\mathbb{R}^{n_\w}\rightarrow\mathbb{R}^{n_g}$, with an uncertain variable $\w$ that relates the uncertainty to the a-priori unknown constraints, and where the uncertainty for each prediction time has a bounded support, i.e., $\wb\in\W_{n|k}\subseteq\mathbb{R}^{n_\w}$. Using $\gamma$, we can then define the constraint $g$ at each prediction time $n$ as
	\begin{equation}
		\label{eq:constr_prediction}
		\gb[n][k](\xb,\ub) := \max_{\wb\in\W_{n|k}} \ \gamma_{n|k}(\xb,\ub,\wb),
	\end{equation}
	which implies robust constraint satisfaction, i.e.,
	\begin{align*}
		\gb[n][k](\xb,\ub) \leq 0 && \Leftrightarrow && \left \{\begin{array}{l}
			\gamma_{n|k}(\xb,\ub,\wb) \leq 0, \\ \forall \ \wb\in\W_{n|k}.
		\end{array}\right.
	\end{align*}
	Furthermore, with a slight abuse of notation, we also denote
	\begin{equation*}
		\gamma_{n|k}(\xb,\ub,\W_{n|k}) := \max_{\wb\in\W_{n|k}}\gamma_{n|k}(\xb,\ub,\wb),
	\end{equation*}
	which will be used to simplify the discussion in Section~\ref{sec:collision_avoidance}.
	
	For our considered setting, the variable $\wb$ represents the state of a dynamical system modeling the future motion of other road users and can be written as
	\begin{equation}\label{eq:obstacle_model}
		\wb[n+1] = \omega(\wb,\boldsymbol{\xi}_{n|k},\xb,\ub),
	\end{equation}
	where $\boldsymbol\xi_{n|k}\in\Xi\subseteq\mathbb{R}^{n_\xi}$ is the associated control variable acting as a source of (bounded) noise. The function~\eqref{eq:obstacle_model} includes an explicit dependence on $\xb$ and $\ub$ in order to model possible interactions between the uncertainty and the system~\eqref{eq:nonlinear_model}. This allows one to, e.g., model the fact that a pedestrian decision might depend on how the vehicle moves. Whether such models can be derived and whether that is desirable or not is beyond the scope of this paper.
	
	\begin{Example}
		In the case where other road users are broadcasting their exact trajectory, function~\eqref{eq:obstacle_model} simply becomes the system dynamics of the other road user, i.e., $\w_{n+1|k} = \omega(\w_{n|k},\boldsymbol{\xi}_{n|k})$.
	\end{Example}

	Furthermore, since~\eqref{eq:obstacle_model} describes how the uncertainty $\w_{n|k}$ evolves in time, it becomes natural to rely on reachability analysis~\cite{althoff2010reachability} to compute outer-approximations of the uncertainty
	\begin{align}\label{eq:reachable}\begin{split}
			\W_{n+1|k}(\xb,\ub) :\supseteq \{\, &\omega(\wb,\boldsymbol\xi_{n},\xb,\ub)\, |\\ & \wb\in\W_{n|k},\ \forall\,\boldsymbol\xi_{n}\in\Xi \, \},
	\end{split}\end{align}
	for some initial $\mathcal{W}_{k|k}=\w_{k|k}$.
	
	To further clarify the nature of the uncertainty sets~\eqref{eq:reachable}, we provide the following example.
	
\begin{figure}[t]
		\centering
		\includegraphics[width=\columnwidth]{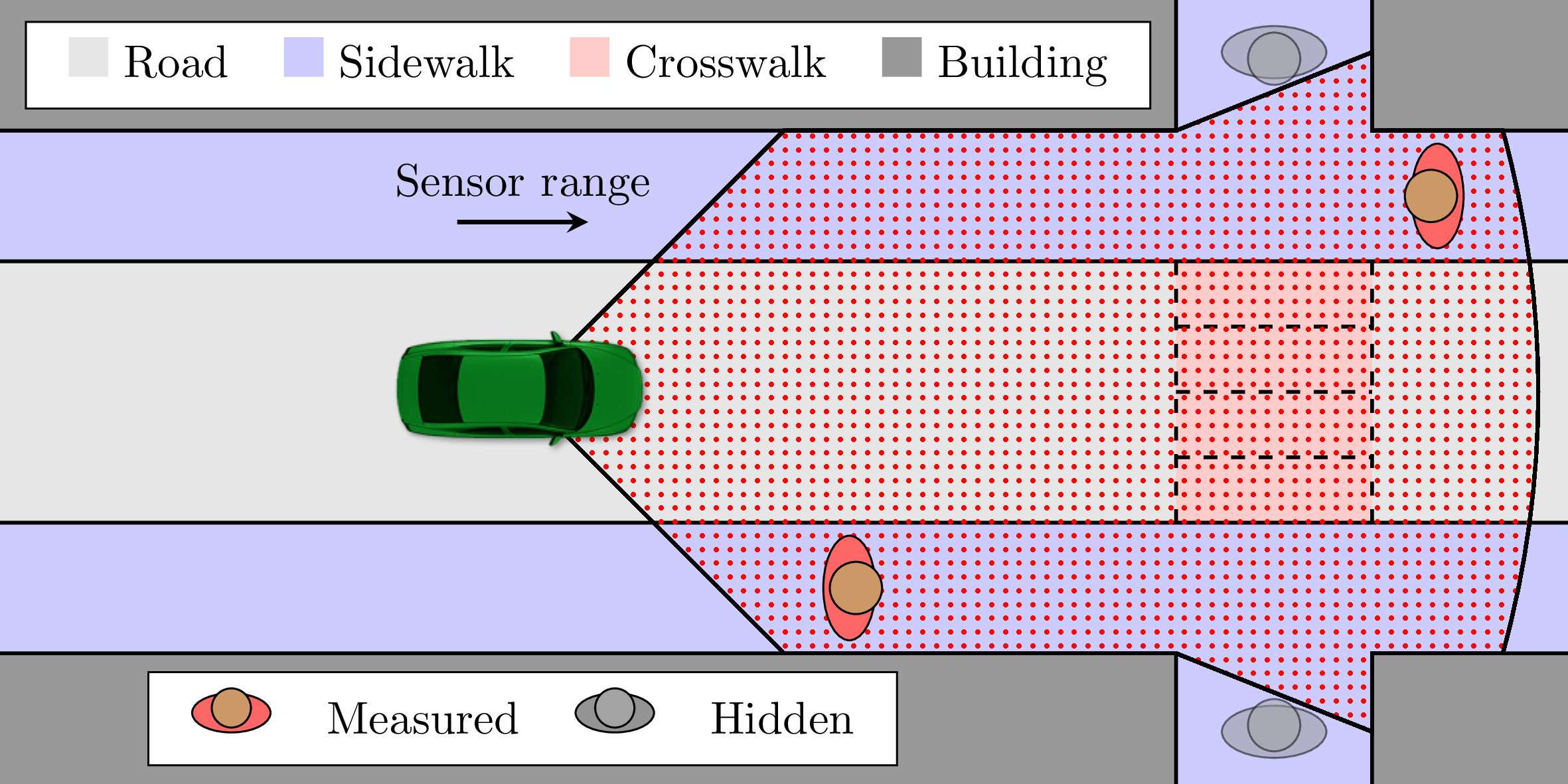}
		\caption{Situation where the sensor range is not able to see pedestrians behind the corner of the building.}\label{fig:occluded}
	\end{figure}

	\begin{Example}\label{ex:agents}
		In a non-cooperative, non-connected, multi-agent setting, the behavior of the other agents is not fully known. In this case, we must distinguish between two types of agents: (a) those that are detectable by the sensors, and (b) those that are either beyond the sensor range or occluded by other obstacles, see e.g., Figure~\ref{fig:occluded}. For type (a) we require that~\eqref{eq:obstacle_model} does not underestimate the set of future states that can be reached by the other agents. For type (b) the uncertainty model must instead account for the possibility that agents could suddenly appear at the sensor range boundary or from behind an obstacle.
	\end{Example}

	The following lemma proves that Assumption~\ref{a:unknown} is satisfied when the uncertainty sets are given by~\eqref{eq:reachable}.
	\begin{Lemma}\label{lem:uncertainty}
		Suppose that $\gb$ is defined according to~\eqref{eq:constr_prediction} with $\mathcal{W}_{n|k}$ satisfying~\eqref{eq:reachable}. Then, Assumption~\ref{a:unknown} holds.
		\begin{proof}
			The proof is given in~\cite[Lemma~1]{batkovic2020safe}
		\end{proof}
	\end{Lemma}
	
	Lemma~\ref{lem:uncertainty} simply states that if the prediction model~\eqref{eq:obstacle_model} is propagated using~\eqref{eq:reachable}, then the uncertainty cannot increase as additional information becomes available (from either the onboard sensors or through communication links). 
	
	In order to simplify the following analysis, we will only consider pedestrian road users in the remainder of this paper. Therefore, we introduce next a pedestrian prediction model that satisfies Assumption~\ref{a:unknown}.
	
	\subsubsection{Pedestrian Prediction Model}

	\begin{figure}[t]
		\centering
		\includegraphics[width=\columnwidth]{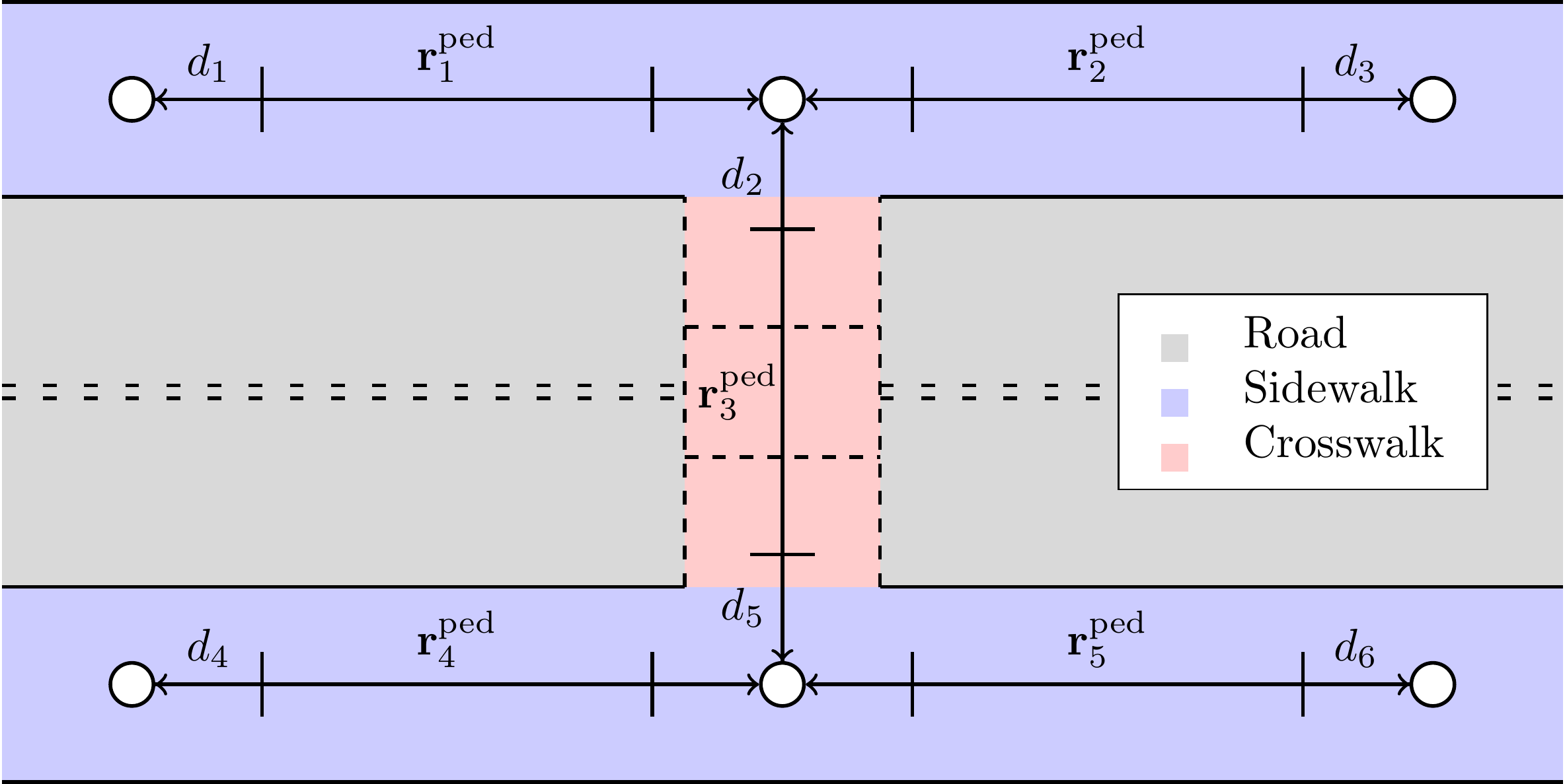}
		\caption{Illustration of different road segments of an intersection can be represented as a connected graph.}\label{fig:ped_map}
	\end{figure}

	In order to predict the future pedestrian motion, we propose a model that leverages the information from the road configuration, e.g., we want to use the information of available sidewalks and crosswalks since pedestrians are expected to use them when in traffic. Such information can be naturally represented as a graph of connected edges, where each edge $i$ is associated with a reference path $\r^{\ped}_i$. An example illustrating this is shown in Figure~\ref{fig:ped_map}.
	
	Having access to such information makes it possible to decompose the pedestrian motion along each path $\r^{\ped}_i$. For this specific setting, we assume that the pedestrian states are given by $\w_k := [w_k^\mathrm{lon},\ w_k^\mathrm{lat}]\in\mathbb{R}^2$ and model the longitudinal and lateral position along path $\r^{\ped}_i$, while $\boldsymbol\zeta_k$ represents the walking speed in both the longitudinal and lateral direction. To contain the uncertainty of the future motion, we place the assumption that the lateral motion will be stabilized, i.e., we place the somewhat strict assumption that pedestrians are rational and abide by the road code. We therefore select the feedback law that guides the prediction to be
	\begin{equation}
		\boldsymbol\zeta_k = [v^{\ped}_i,\ -K w_k^{\mathrm{lat}}],
	\end{equation}
	where $v^\mathrm{ped}_i$ denotes the reference walking speed along path $\r^{\ped}_i$, and $K>0$ is a feedback gain. With this state representation, we can write the closed-loop pedestrian model as
	\begin{align}\label{eq:ped_model}
		\begin{split}\w_{k+1}&=\omega_i(\w_k,\boldsymbol\xi_k)\\&=\matr{cc}{1&0\\0&1-t_\mathrm{s}K}\w_k + \matr{c}{t_\mathrm{s}\\0}v^{\ped}_i+\matr{cc}{t_\mathrm{s}&\hspace{-0.5em}0\\0&\hspace{-0.5em}t_\mathrm{s}}\boldsymbol\xi_k,\end{split}
	\end{align}
	where $\boldsymbol\xi\in\Xi\subseteq\mathbb{R}^{2}$, i.e., $\|\boldsymbol\xi\|_\infty\leq\bar{\xi}$, is assumed to be some bounded noise with zero mean, representing some uncertainty in the future motion. The pedestrian motion can then be predicted using
	\begin{align}\label{eq:nom_ped_model}
		\w^\mathrm{nom}_{n+1|k}&=\matr{cc}{1&0\\0&1-t_\mathrm{s}K}\w^\mathrm{nom}_{n|k} + \matr{c}{t_\mathrm{s}\\0}r^{\ped,i}_v,
	\end{align}
	for an initial $\w^\mathrm{nom}_{k|k}=\w_k$. Finally, by using reachability analysis~\cite{althoff2010reachability,mulagaleti2021computation}, we can also predict the uncertainty sets $\mathcal{W}_{n|k}$ along an edge $i$ similarly to~\eqref{eq:reachable}. As the longitudinal and lateral dynamics are decoupled for our proposed model, we can easily propagate the uncertainty in each direction, i.e., one can construct the uncertainty set at the next time step as
	\begin{align}
		\begin{split}
			\W_{n+1|k} := \Bigg\{\w\ |\ 
			&\w\leq \matr{c}{
				w_{n|k}^\mathrm{lon}+t_\mathrm{s}(v_i^\mathrm{ped}+\bar{\xi})\\
				(1-t_\mathrm{s}K)w_{n|k}^\mathrm{lat}+t_\mathrm{s}\bar{\xi}},\\
			& \w \geq \matr{c}{
				w_{n|k}^\mathrm{lon}+t_\mathrm{s}(v_i^\mathrm{ped}-\bar{\xi})\\
				(1-t_\mathrm{s}K)w_{n|k}^\mathrm{lat}-t_\mathrm{s}\bar{\xi}
			},\\
			& \forall\  \wb:=[w_{n|k}^\mathrm{lon},w_{n|k}^\mathrm{lat}]^\top\in\W_{n|k}
			\Bigg\},
		\end{split}
	\end{align}
	with $\W_{k|k}=\w_k$. More complex models can be straightforwardly accommodated by applying standard computational geometry tools~\cite{Kolmanovsky1998,borrelli2017predictive,MPT3}.

	Note that~\eqref{eq:ped_model}  describes the predicted pedestrian motion along edge $i$ only. Since the road map in Figure~\ref{fig:ped_map} consists of a set of connected edges, the prediction will need to transition between the different edges eventually. To that end, similarly to~\cite{batkovic2018computationally}, we transition the dynamics from edge $i$ to the neighboring connected edges, whenever the nominal prediction $\w_{n|k}^\mathrm{nom}$ is within a transitioning threshold, e.g., whenever the nominal prediction is at a certain distance from the final node. Note that, if the final node is connected to multiple edges, then the future predictions are propagated along all these edges. This enables the model to account for all possible directions that a pedestrian might take, making it possible to consider multi-modal predictions.
	
	In reality, pedestrian walking behaviors may differ depending on numerous factors, e.g., personal preferences, the environment, and the presence of other pedestrians, which makes the future movement difficult to model in detail. While we proposed this pedestrian model for simplicity, we must stress that other frameworks/models can also be used as long as Assumption~\ref{a:unknown} is satisfied. In particular, the recently proposed methods in~\cite{koschi2018set,koschi2020set}, where the authors use map-based information and incorporate traffic rules in their predictions could also be applied.
	
	Next, we show how the a-priori unknown constraint can finally be expressed using the derived prediction model~\eqref{eq:ped_model}.

	\subsubsection{Constraint generation}
	
	\begin{figure}[t]
		\centering
		\includegraphics[width=\columnwidth]{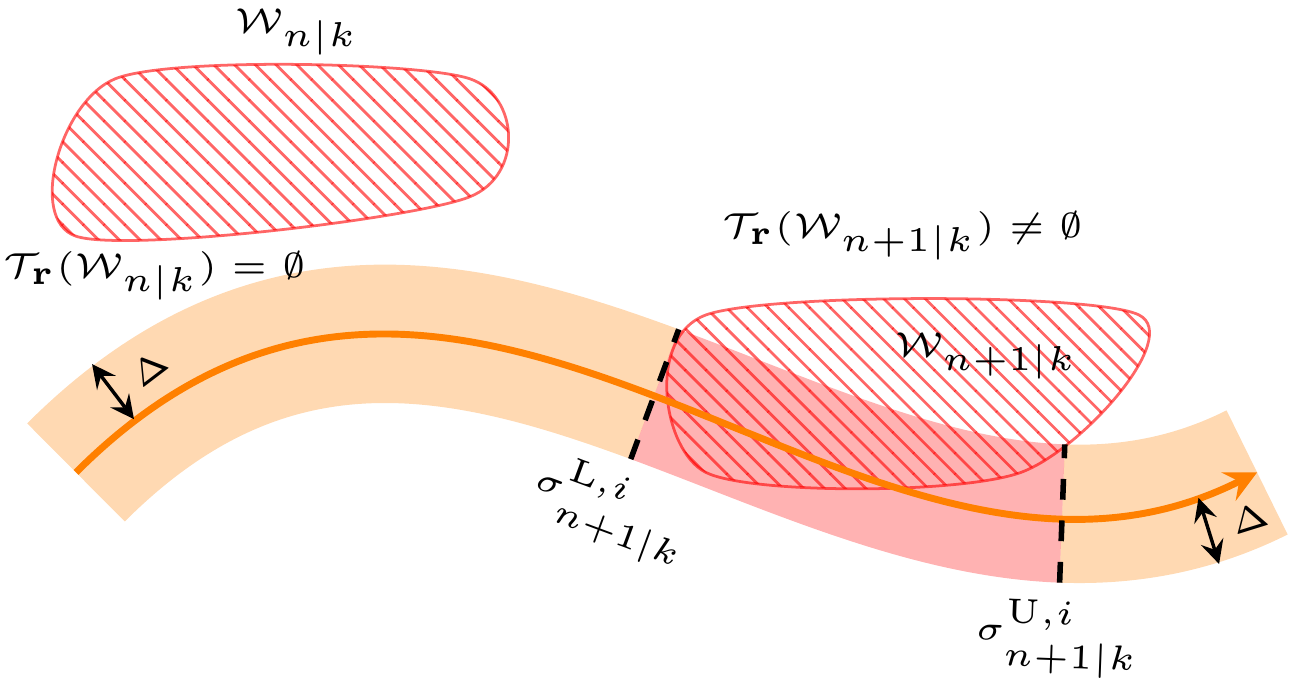}
		\caption{A visual representation of~\eqref{eq:reference_tube} and~\eqref{eq:collision_set}. The orange line represents the vehicle's reference path $\rx(s)$, while the red regions represent the uncertainty sets $\W_{n|k}$ and $\W_{n+1|k}$.}\label{fig:clarify_d_sigma}
	\end{figure}

	For each road user $i$ that is present in the environment, we need to identify if the vehicle potentially needs to interact with it, e.g., if the predicted path of the road user eventually crosses the planned path of the vehicle. Hence, having the uncertainty set $\W_{n|k}^i$ for each road user $i$, we want to avoid the positions where the uncertainty enters within a distance $\Delta:=\bar{e}_y+\Delta_\mathrm{safe}$ of the reference trajectory, where $\Delta_\mathrm{safe}$ accounts for the dimensions of the ego vehicle, as well the obstacle dimensions. More formally, the longitudinal positions that should be avoided can be expressed through the following set
	\begin{align}
		\begin{split}\label{eq:reference_tube}
			\mathcal{T}_\r(\W_{n|k}) := \{ s\ | \
			&\exists \, \wb\in\W_{n|k}\ \text{s.t.}\\
			& \|T^\w(\wb)-T^\x(\rx(s))\|_2\leq \Delta \},
		\end{split}
	\end{align}
	where $T^\x:\mathbb{R}^{n_\x}\rightarrow\mathbb{R}^2$ and $T^\w:\mathbb{R}^{n_\w}\rightarrow\mathbb{R}^2$ map the position components of the reference and the uncertainty set into a common global frame. A visual example of set~\eqref{eq:reference_tube} is shown in Figure~\ref{fig:clarify_d_sigma}. Using~\eqref{eq:reference_tube} we can then express the collision avoidance set for each road user $i$ as
	\begin{equation}\label{eq:collision_set}
		\mathcal{C}_{n|k}^i(\mathcal{W}_{n|k}^i) := \begin{cases}
			[\sigma_{n|k}^{\mathrm{L},i},\ \bar\sigma_{n|k}^{\mathrm{U},i}]&\text{if}\ \mathcal{T}_\r(\W_{n|k}^i)\neq\emptyset,\\
			\emptyset, &\text{otherwise},
		\end{cases}
	\end{equation}
	where we define $\sigma_{n|k}^{\mathrm{L},i}$ as 
	\begin{align*}
		\sigma_{n|k}^{\mathrm{L},i} := \min_{s} \ & s &&\mathrm{s.t.} \ s\in\mathcal{T}_\r(\mathcal{W}_{n|k}^i),
	\end{align*}
	and $\sigma_{n|k}^{\mathrm{U},i}$ as the solution of the same problem with the minimization replaced by a maximization. To further clarify the meaning of the collision avoidance set $\mathcal{C}_{n|k}^i$, we provide the following example.

	\begin{figure}[t]
		\centering
		\includegraphics[width=\columnwidth]{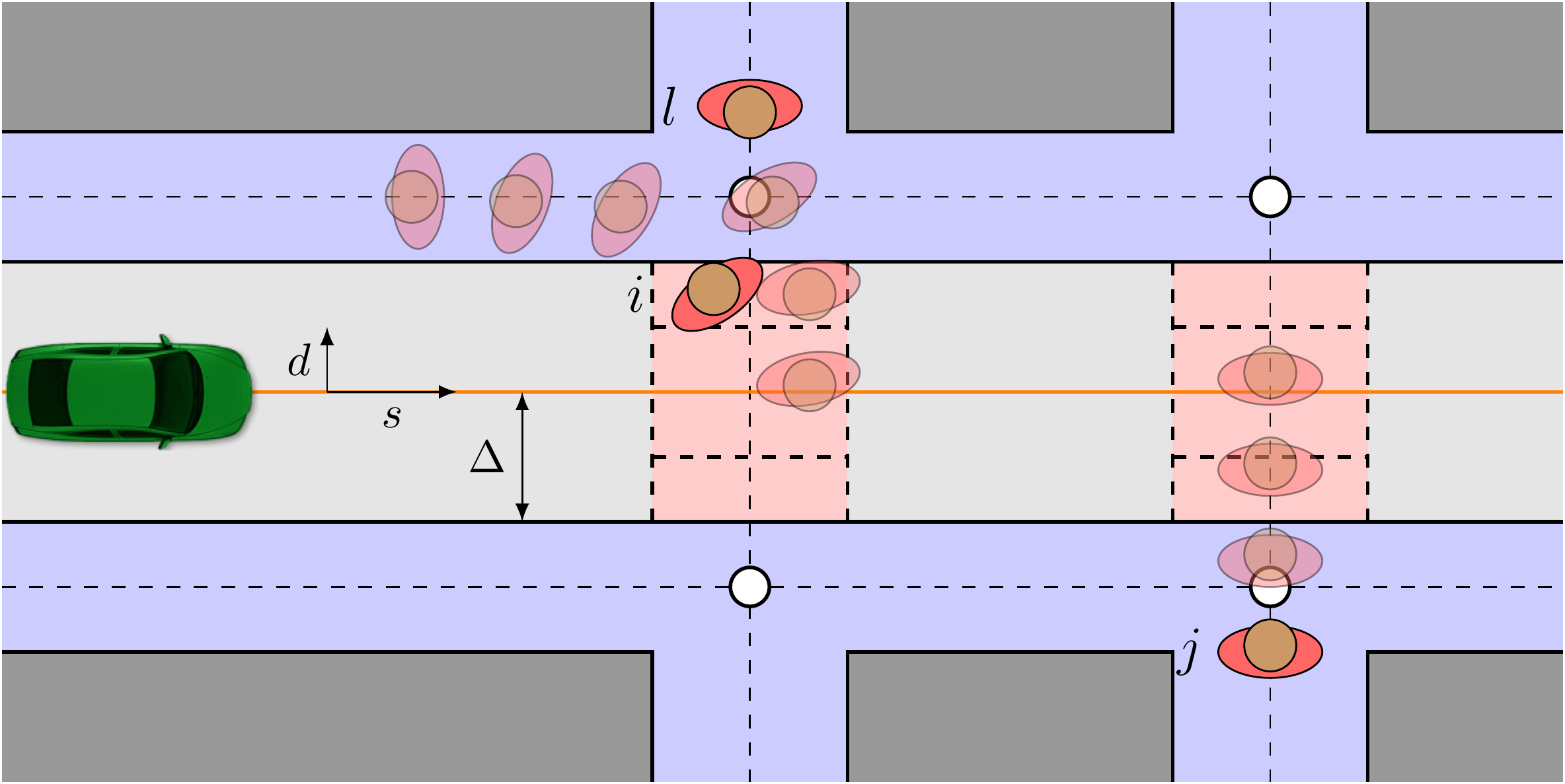}
		\caption{The orange line represents the vehicle's reference path, while the black dashed lines represent the walkable road graph of the pedestrian model.}\label{fig:peds}
	\end{figure}

\begin{figure}[t]
		\centering
		\includegraphics[width=\columnwidth]{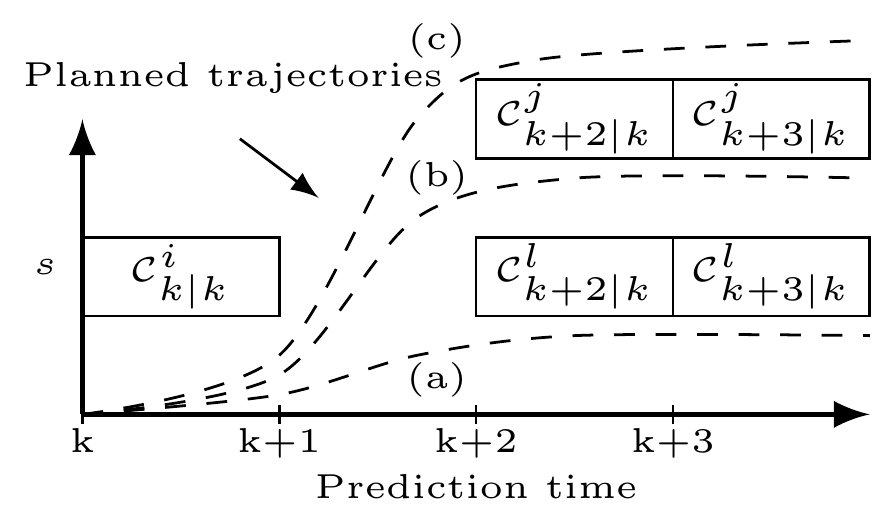}
		\caption{The collision avoidance sets $C_{n|k}^z$, $z\in\{i,j,l\}$ for the three pedestrians from Example~\ref{ex:collision-avoidance}. The boxes represent the longitudinal positions which the vehicle must avoid, while the dashed lines represent three different possible planned trajectories the vehicle can make in this setting.}\label{fig:collision_areas}
	\end{figure}

	\begin{Example}\label{ex:collision-avoidance}
		Consider the setting presented in Figure~\ref{fig:peds}, where the future motions of three pedestrians are predicted. In this situation, pedestrian $i$ is only within the driving corridor $\Delta$ at the initial time, hence, $\mathcal{C}_{k|k}^i\neq\emptyset$ and $\mathcal{C}_{k'|k}^i=\emptyset$ for all $k'>k$, while the other two pedestrians $i$ and $j$ are instead predicted to enter the driving corridor at a later time. In this particular setting, the vehicle has three options to consider: (a) stop for all pedestrians, (b) yield to pedestrian $i$ and $j$, or (c) yield only to pedestrian $i$. Figure~\ref{fig:collision_areas} provides an illustration of the collision avoidance sets and shows the three possible decisions that the vehicle can take, while avoiding collisions.
	\end{Example}
	
	To formulate constraint $g_{n|k}$ we realize that the vehicle must decide between either driving ahead of the road user, or if it should yield to it. To that end, for each road user $i$, we must avoid the positions contained within the collision-avoidance set $\mathcal{C}_{n|k}^i$, and therefore formulate the following candidate collision avoidance function
	\begin{align}\label{eq:gamma}
		&\gamma^i_{n|k}(s_{n|k},\ub,\W_{n|k}^i) :=\\ 
		&\begin{cases}
			s_{n|k} - \sigma_{n|k}^{\mathrm{U},i} &\hspace{-.5em} \text{if yielding and}\ \mathcal{C}_{n|k}^i(\W_{n|k}^i)\neq\emptyset,\\
			\sigma_{n|k}^{\mathrm{L},i}-s_{n|k} &\hspace{-.5em} \text{if not yielding and}\ \mathcal{C}_{n|k}^i(\W_{n|k}^i)\neq\emptyset,\\
			0 &\hspace{-.5em} \text{otherwise.}
		\end{cases}\nonumber
	\end{align}
	which finally allows us to write the resulting a-priori unknown constraint as
	\begin{align}\label{eq:g_constr}
		g_{n|k}^i(s_{n|k},\ub) := \gamma_{n|k}^i(s_{n|k},\ub,\W_{n|k}^i).
	\end{align}
	
	\begin{Remark}\label{remark:combo}
		While the introduction of additional pedestrians could be challenging, one could use some strategy to reduce the complexity while still capturing the most important features of the problem. Note, however, that only a limited amount of pass/yield configurations can effectively exist in the prediction horizon, e.g., if too many pedestrians try to cross the road, the only feasible option might be to stop and wait for the them to pass.
	\end{Remark}
	
	As a safe though possibly conservative approach to address the combinatorial complexity issue that Remark~\ref{remark:combo} mentions, we deploy a simple heuristic in this paper that only considers the combinations of the closest intersection, and always yield to road users further away. In particular, for the closest intersection, we sort the times when the road users enter and solve Problem~\eqref{eq:nmpc} for the different combinations. Considering the setting presented in Example~\ref{ex:collision-avoidance} and Figures~\ref{fig:peds} and \ref{fig:collision_areas}, this amounts to considering the options: (a) yield to both road users $i$ and $l$, and (b) yield to road users $i$ and $j$.
	
	Finally, as it was mentioned in Example~\ref{ex:agents}, in order to ensure the validity of Assumption~\ref{a:unknown}, the vehicle must also be ready to act for any road users that cannot be measured directly from its sensors, i.e., it must account for the possibility that other road users may be occluded similarly to Figure~\ref{fig:occluded}. To address this issue, we deploy a similar strategy as in~\cite{koschi2020set}, where virtual road users are placed wherever the sensor-suite warns that a potential occlusion may occur. Then, for each virtual road user, we predict the future uncertainty sets $\mathcal{W}^\mathrm{virtual}_{n|k}$ and form constraints~\eqref{eq:g_constr}.

	While Sections~\ref{sec:system} and~\ref{sec:collision_avoidance} showed how the known and unknown constraints could be constructed for the vehicle model~\eqref{eq:nonlinear_model}, the final step needed  for Proposition~\ref{prop:recursive} is to show how the terminal conditions can be derived.

	\subsection{Safe Terminal Conditions}\label{sec:terminal}

	This section addresses the derivation of the stabilizing terminal set needed for Assumption~\ref{a:stab} and provides an example of a safe set that is suitable for urban autonomous driving and satisfies Assumption~\ref{a:safe}.
	
	\subsubsection{Stabilizing Terminal Conditions}
	
	In order to derive the terminal cost matrix $P$, and the stabilizing terminal set $\mathcal{X}_\r^\mathrm{s}$ for Problem~\eqref{eq:nmpc} we propose to first decouple the longitudinal and lateral motions of the nonlinear system~\eqref{eq:error_model}. While the full derivation is provided in Appendix~\ref{appendix:A}, here we will only provide an overview of the steps for convenience. Hence, the reader is referred to the Appendix for the full details of the derivation.
	
	Since the longitudinal kinematics of~\eqref{eq:error_model} are linear, i.e., the states $\x_\mathrm{lon}:=[v,\ a]^\top$, we use standard approaches in the literature to compute the stabilizing terminal conditions~\cite{blanchini2008set}. In particular, we use an LQR controller with tuning $Q^\mathrm{LQR}_\mathrm{lon}\in\mathbb{R}^{2\times2}$ and $R^\mathrm{LQR}_\mathrm{lon}\in\mathbb{R}$ to obtain a feedback gain $K_\mathrm{lon}$ that stabilizes the system and then solve the Linear Matrix Inequality~(LMI)~\eqref{eq:long_lmi} to obtain the longitudinal terminal cost matrix $P^\mathrm{lon}$. To get the corresponding longitudinal terminal set $\mathcal{X}_\r^\mathrm{lon}$ we start from the admissible constraint set, and iteratively compute backwards reachable sets that are intersected with the admissible constraint set~\cite{Kolmanovsky1998}. Repeating this process until convergence yields the following longitudinal terminal invaraiant set
	\begin{equation}
		\mathcal{X}^\mathrm{lon}_\r(s):=\left \{v,a\ |\ H_\mathrm{lon}\matr{c}{v-v^\r(s)\\ a-a^\r(s)}\leq b_\mathrm{lon}\right \}.
	\end{equation}

	The lateral kinematics of~\eqref{eq:error_model}, with the corresponding states $\x_\mathrm{lat}:=[e_y,\  e_\psi,\ \delta,\ \alpha]^\top$, can be viewed as a Linear Parameter Varying~(LPV) system whose dynamics depend on the velocity. As such, we construct a polytopic approximation of the nonlinear system for a range of velocities $v$~\cite{boyd1994linear}. Then, similarly to the longitudinal kinematics, we solve LMI~\eqref{eq:lat_lmi} to obtain the terminal cost matrix and compute the lateral terminal set
	\begin{equation*}
		\mathcal{X}_\r^\mathrm{lat}(s) = \{ e_y,e_\psi,\delta,\alpha\ |\ H_\mathrm{lat}[e_y,e_\psi,\delta-\delta^\r,\alpha-\alpha^\r]^\top\hspace{-0.4em}\leq{}b_\mathrm{lat}\}.
	\end{equation*}
	
	The terminal cost matrix $P$ and stabilizing terminal set $\mathcal{X}_\r^\mathrm{s}$ are then finally expressed as
	\begin{align}
		P&=\mathrm{blockdiag}(P_\mathrm{lat},P_\mathrm{lon})\\
		\mathcal{X}_\r^\mathrm{s}(s) &:= \{\x\ |\  [v,\ a]^\top\hspace{-0.5em}\in\mathcal{X}_\r^\mathrm{lon}(s), [e_y,\ e_\psi,\ \delta,\ \alpha]^\top\hspace{-0.5em}\in\mathcal{X}_\r^\mathrm{lat}(s)\}.\nonumber
	\end{align}
	We remind the reader that the full derivation of the terminal conditions is presented in Appendix~\ref{appendix:A}.

	\begin{figure}
		\centering
		\includegraphics[width=\columnwidth]{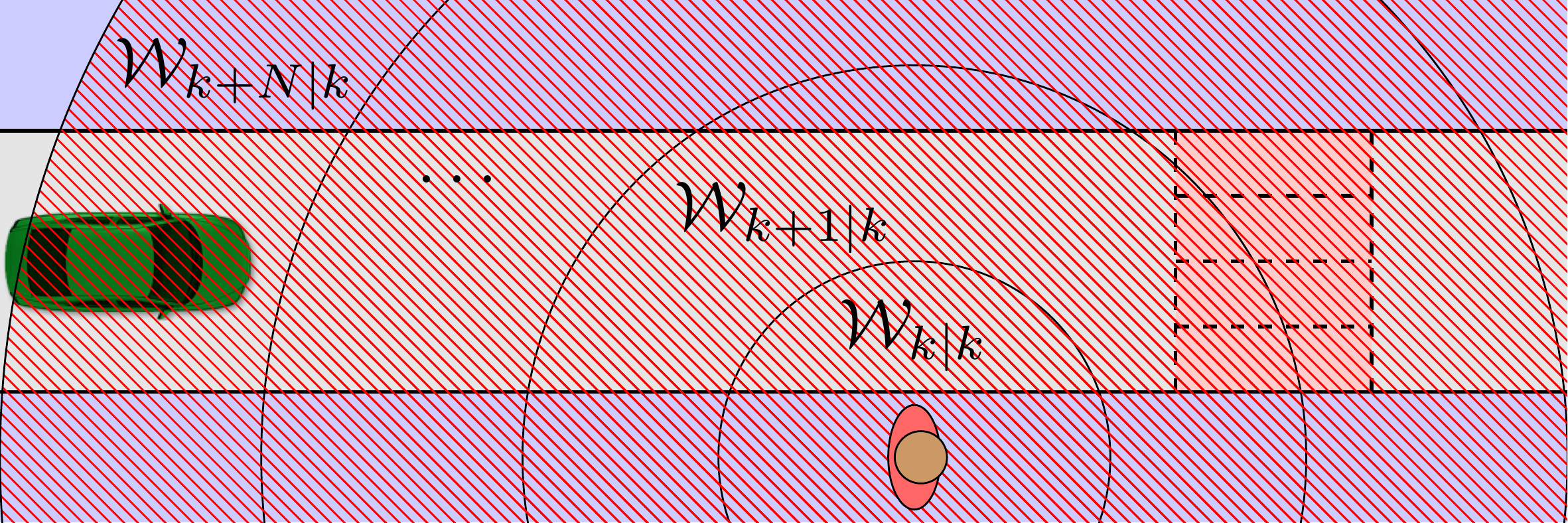}
		\caption{As the predicted uncertainty $\mathcal{W}_{n|k}$ grows in time, the collision-free region of the vehicle drastically shrinks.}\label{fig:uncertainty-example}
	\end{figure}
	
	\subsubsection{Safe Set Formulation}
	For many practical applications where safety is emphasized, a system is generally considered to be safe at steady state, i.e., 
	\begin{align}\label{eq:safe_set}
		\begin{split}
			\mathcal{X}_\mathrm{safe}(s_k):= \{\x\ |\ &\x=f(\x,\u),\\&h_k(\x,\u)\leq{}0,\ m_k(\x,\u)\leq{}0  \},
		\end{split}
	\end{align}
	where the function $m_k$ may define additional constraints needed for the set definition. This general definition can readily be applied to autonomous driving settings, since a vehicle is generally considered safe if it has come to a full stop, i.e., a vehicle that is parked in a parking lot, or other safe configurations modeled by $m_k$ are considered safe and not responsible for collisions with other road users. While the safe set in~\eqref{eq:safe_set} serves as an example in this paper, we must stress that other formulations of the safe set are possible. For instance, one could replace~\eqref{eq:safe_set} with a time-to-collision constraint set, provided that one is able to guarantee that this does yield the required safety guarantees.
	
	In general, most processes controlled either by humans or by automatic controllers do have emergency procedures which are triggered whenever safety is jeopardized. Assumption~\ref{a:safe}, i.e., the existence of a safe set, is meant to cover all these situations. Furthermore, we must stress that this assumption also entails that $\gb$ directly depends on the safe set, since it is assumed that $\gb$ may never be active when $\xb\in\mathcal{X}_\mathrm{safe}(s_n)$. Hence,
	this assumption entails that when using the uncertainty model presented in Section~\ref{sec:collision_avoidance}, function $\gamma$ used in~\eqref{eq:gamma} is not simply the output of the uncertainty model, but also includes the information that $\xb\in\mathcal{X}_\mathrm{safe}(s_{n|k})$ implies $\gamma_{n|k}(\xb,\u_\mathrm{safe},\wb) = 0$.

	To further motivate the choice of a safe set at steady state, consider the setting presented in Figure~\ref{fig:uncertainty-example}, where a prediction of the future pedestrian motion (red shaded region) is growing unbounded in time. For such a setting, one can argue that the safest option that the vehicle can take is to come a complete stop, since the uncertainty allows the future pedestrian position to be anywhere on the road. However, this choice does not entail that the pedestrian will never hit the vehicle, since the model still predicts that this can happen. The second part of Assumption~\ref{a:safe} is meant to cover this case by imposing that $\xb\in\mathcal{X}_\mathrm{safe}(s_{n|k})$ implies $\gamma_{n|k}(\xb,\u_\mathrm{safe},\wb) = 0$, which entails that the model predicts that the pedestrian will avoid hitting the vehicle if it is stopped. This observation makes the assumption more acceptable, as it yields a more refined pedestrian model which accounts for the fact that pedestrians do not walk over stopped vehicles. Additionally, this is also abiding by the standards of the road code, for which a parked vehicle is not responsible for collisions that might occur. Clearly, more complex situations (in terms of responsibilities in case of collisions) arise when the vehicle needs to stop in a crossroad in order to yield to another vehicle having priority, or at a zebra crossing to yield to a pedestrian. Covering all possible situations in the best way is beyond the scope of this paper and the subject of future research.
	
	In this paper, we express the safe set as all states where the vehicle has come to a full stop.  We motivate this choice by noting that human drivers, in general, already plan their driving such that they can safely come to a complete stop in case of an emergency situation, i.e., reaching a zero velocity is considered to be safe. While what we propose here might be over-simplified for real-world autonomous driving applications, future work will carefully evaluate whether more refined strategies are necessary. Note that other formulations of the safe set might be better suited for settings outside the scope of urban driving, e.g., for highway driving one could instead rely on a safe set that considers a specific relative distance towards the lead vehicle while keeping a zero difference in velocity.

	We now have all the necessary results needed to formulate~\eqref{eq:nmpc} and implement the theory from Proposition~\ref{prop:recursive}. In the next section, we deploy the controller based on Problem~\eqref{eq:nmpc} and show the importance of Assumption~\ref{a:unknown}, and how recursive feasibility (safety) can be jeopardized if occluded obstacles are not considered. Then, in Section~\ref{sec:experiments} we deploy the framework in a real vehicle at a test track to verify the real-time performance of the controller.

	\section{Simulation Results}\label{sec:simulation}

	\begin{table} \caption{Parameter values for Sections~\ref{sec:simulation} and~\ref{sec:experiments}.}
		\centering
		\begin{tabular}{|lll||lll|}
			\hline
			$w_0$ &  $20$ & [s$^{-1}$] & $w_1$ & $0.9$ 	& [-]\\
			$t_\mathrm{acc}$ &  $1.8$ & [s$^{-1}$] & $\bar{e}_y$ & $0.4$ 	& [m]\\
			$\bar{e}_\psi$ &  $0.61$ & [rad] & $\bar{\delta}$ & $0.53$ 	& [rad]\\
			$\bar{v}$ & $15.28$ & [m$\cdot$s$^{-1}$] & $\underline{a}$ & $-5$ & [m$\cdot$s$^{-2}$]\\
			$\bar{a}$ & $2$ & [m$\cdot$s$^{-2}$] & $\bar{\alpha}$ & $0.35$ & [rad$\cdot$s$^{-1}$]\\
			\hline
		\end{tabular}
		\label{tab:param}
	\end{table}
	
\begin{figure}[t]
		\includegraphics[width=\columnwidth]{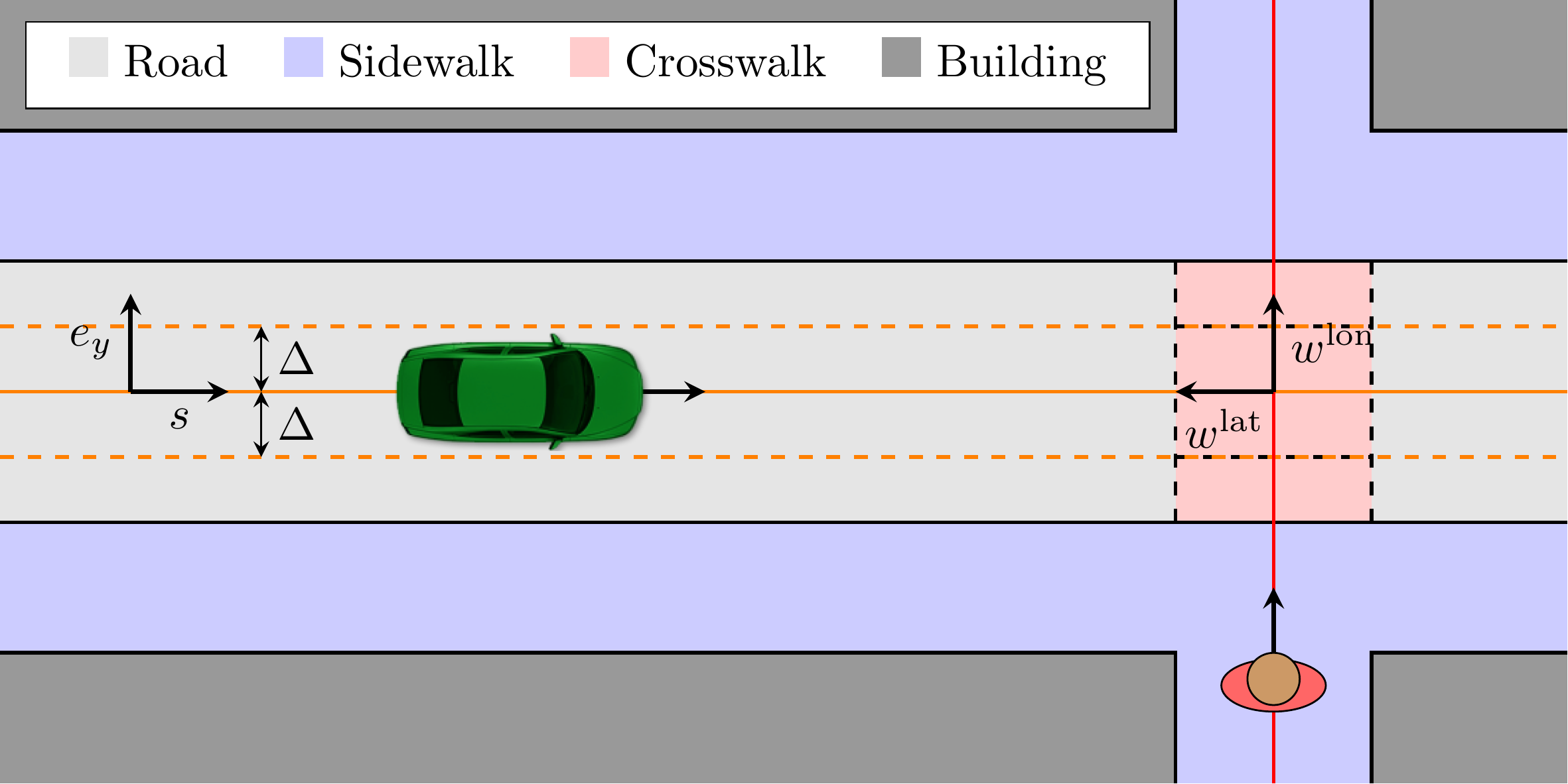}
		\caption{Straight road simulation setting with a crossing pedestrian that is occluded by the environment.}\label{fig:exp-A}
	\end{figure}

	In this section we consider the straight road setting presented in Figure~\ref{fig:exp-A} in order to illustrate how the controller based on  Problem~\eqref{eq:nmpc} behaves in a situation when a pedestrian crosses the road. While this setting is indeed very simple, we will show that an occluded pedestrian that might appear from the bottom in Figure~\ref{fig:exp-A} may render the controller unsafe if the assumptions needed for Proposition~\ref{prop:recursive} do not hold. Therefore, to illustrate the implications that Proposition~\ref{prop:recursive} has on safety, we will show two simulations: one where Proposition~\ref{prop:recursive} does not hold (Assumption~\ref{a:unknown} is not satisfied); and one where Proposition~\ref{prop:recursive} does hold. 
	
	For both simulations we consider the parameter values presented in Table~\ref{tab:param} and we formulate the MPC controller~\eqref{eq:nmpc} in the multiple shooting framework using a sampling time of $t_\mathrm{s}=0.05\ \mathrm{s}$, prediction horizons $N=20$ and $M=100$. We used the following stage cost matrices
	\begin{equation}\label{eq:cost_def}
		Q=\mathrm{diag}(1,1,10,1,1,1),\ R=\mathrm{diag}(4,10),
	\end{equation}
	and the terminal matrix $P$ from~\eqref{eq:terminal_cost_matrix}. The discretized dynamics of~\eqref{eq:error_model} were obtained using a fourth order Runge-Kutta integrator with 5 steps per control interval. The OCP was then solved using Acados~\cite{Verschueren2019} together with the HPIPM solver~\cite{frison2020hpipm} by deploying a Sequential Quadratic Programming~(SQP) approach with a Real Time Iteration~(RTI) scheme~\cite{gros2020linear,diehl2005real}.
	
	\begin{figure}[t]
		\centering
		\includegraphics[width=\columnwidth]{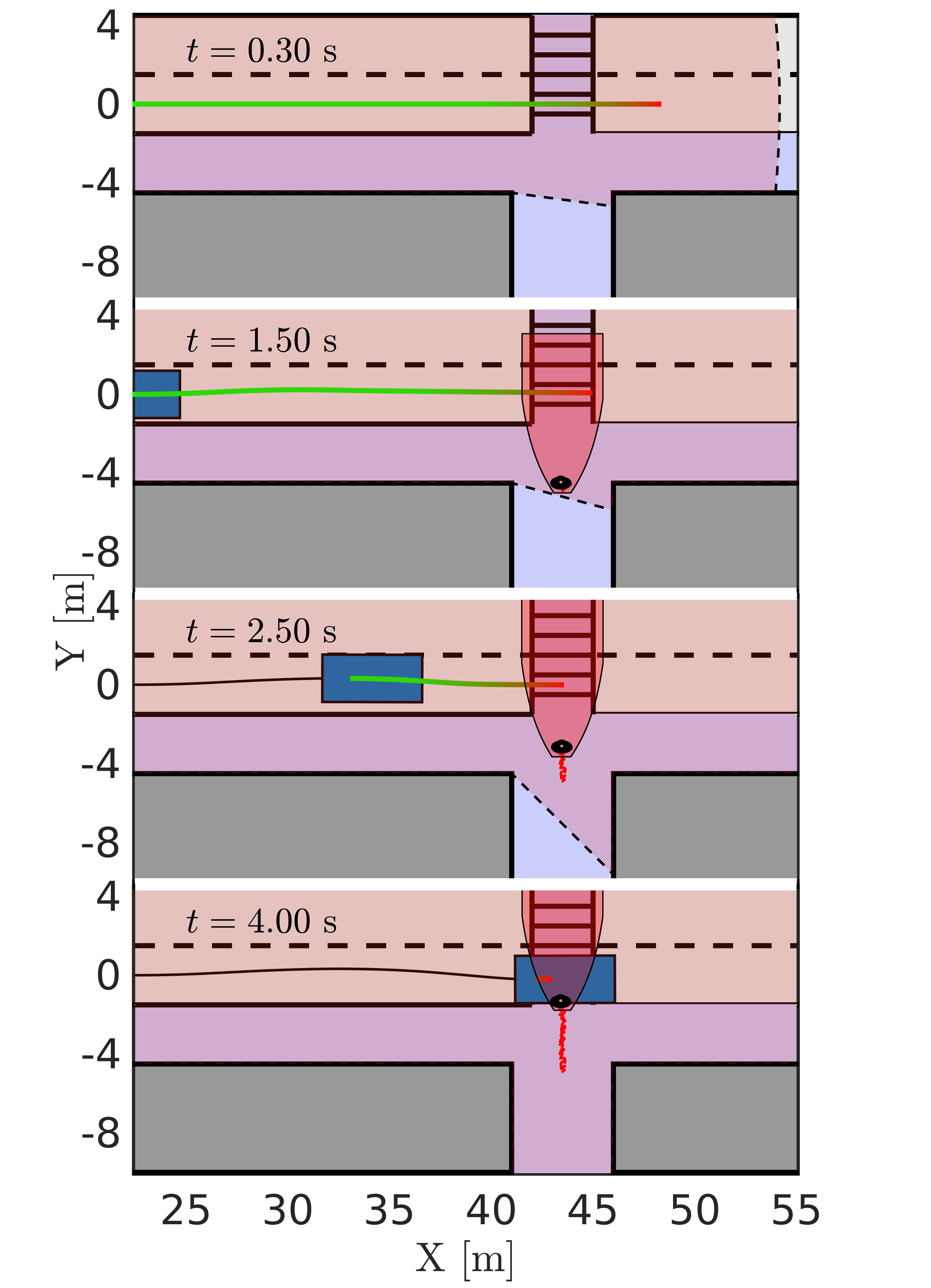}
		\caption{Four different time instances of the simulation environment. The first panel shows that the sensors (pink shaded region) cannot see behind a wall, and that the vehicle therefore plans a trajectory (red/green line indicating low/high speed) within the sensing range. The last three panels show that a pedestrian, who was not visible for the sensors earlier, appears and forces the vehicle to perform an emergency braking, which results in a collision.}
		\label{fig:crash}
	\end{figure} 
	
	\subsection{Compromised Safety}\label{sec:simulation:unsafe}
\begin{figure}[t]
		\centering
		\includegraphics[width=\columnwidth]{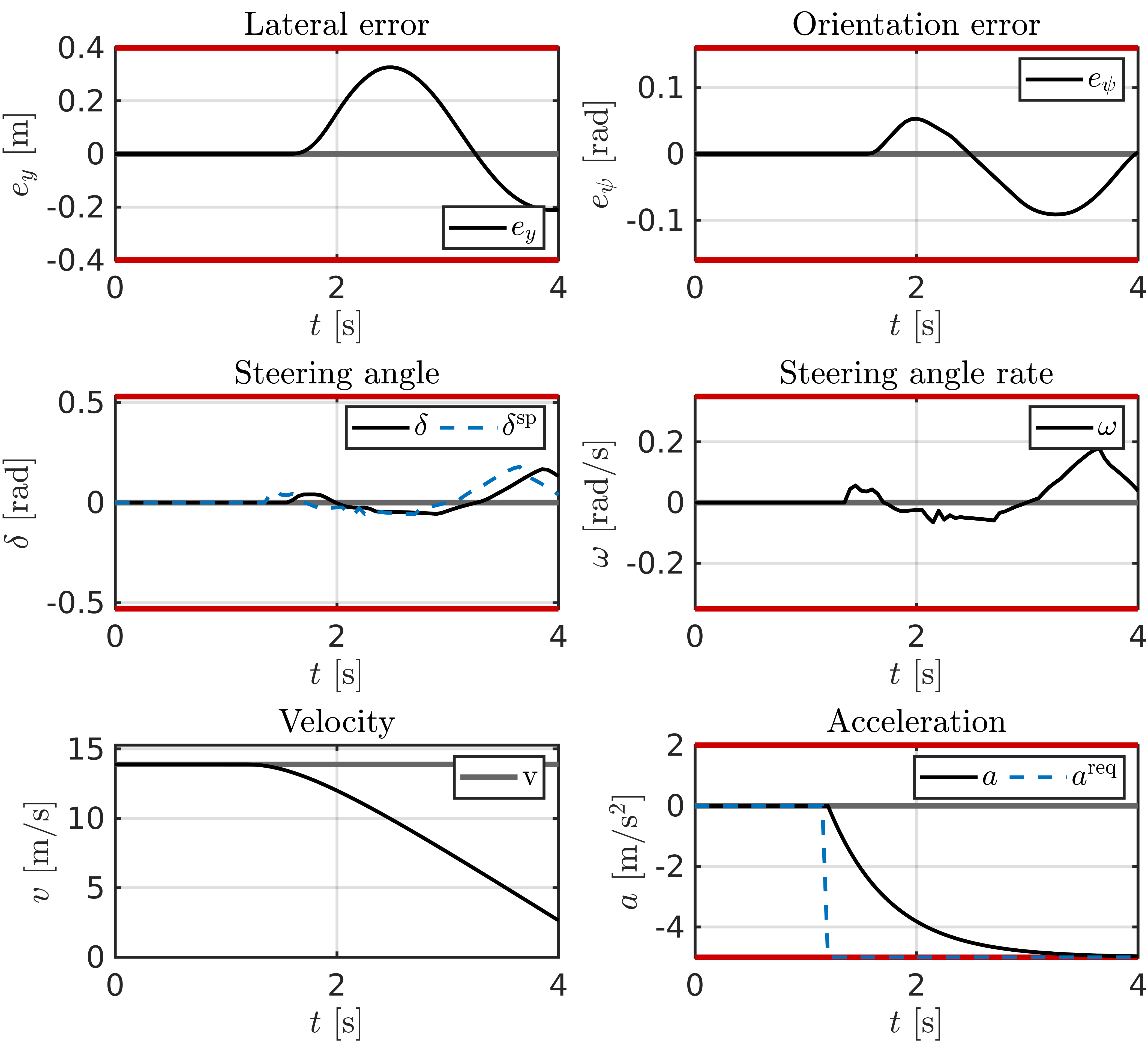}
		\caption{Closed-loop evolution of the unsafe MPC controller shown in Figure~\ref{fig:crash}. Just before $t\leq{}2\ \si{s}$, the pedestrian becomes visible and forces the vehicle to perform an emergency braking. The red lines denote the state and input bounds, while the gray line denotes the reference trajectory.}
		\label{fig:crash_closed}
	\end{figure}

	Figure~\ref{fig:crash} shows four time instances of the MPC controller that only reacts to what it can sense directly, i.e., it does not expect that an occluded road user may appear at the intersection. To that end, it is visible from the first time instance in Figure~\ref{fig:crash} that the vehicle assumes that it can pass the intersection without causing any collision. Around time $t=1.5\ \si{s}$ the occluded pedestrian is eventually detected by the sensor system, and since the vehicle cannot clear the intersection safely, it applies full braking. However, due to the late detection of the road user, the vehicle does not manage to come to a full stop before colliding with the road user. Figure~\ref{fig:crash_closed} shows the closed loop states of the controller which shows that around $t=1.5\ \si{s}$ the vehicle applies full braking and tries to steer to marginally increase the traveled distance before colliding with the pedestrian. While this behavior may not be desirable in practice, it can be addressed by modifying the vehicle model and/or system constraints, e.g., through the inclusion of tire force constraints. Note that even at time $t=4\ \si{s}$ (some time after the initial collision), the velocity is still around $3\ \si{m/s}$ ($\approx10\ \si{km/h}$).

\begin{figure}[t]
		\centering
		\includegraphics[width=.9\columnwidth]{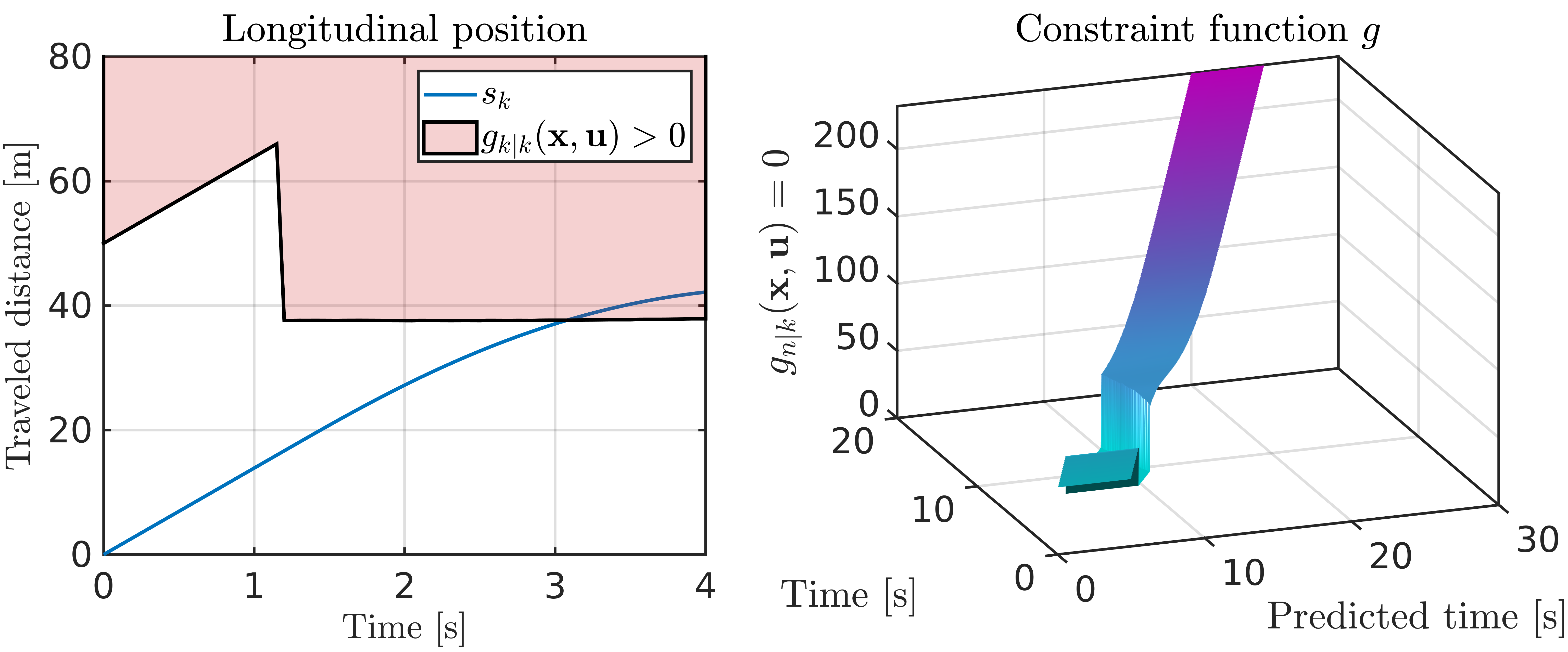}
		\caption{Time evolution of $g_{n|k}$ for the unsafe MPC controller in Section~\ref{sec:simulation:unsafe}. The left-hand side shows the constraint $g_{k|k}$, while the right-hand side shows the predicted $g_{n|k}$.}
		\label{fig:crash_g}
	\end{figure}

	What caused the controller to become unsafe in this simulation setting is the fact that Assumption~\ref{a:unknown} was not satisfied. Figure~\ref{fig:crash_g} shows how $g_{k|k}$ and $g_{n|k}$ evolves in time. From the left plot it is visible that the upper bound on the longitudinal position (red region) drastically shrinks around $t=1.5\ \si{s}$. This is also visible in the right plot where the predicted constraint does not satisfy $g_{n|k+1}\leq g_{n|k}$. The feasibility of the problem was preserved by relaxing constraint $\gb$ with an exact penalty~\cite{scokaert1999feasibility}.
	
	Next, by considering the same example, we show that safety does not become jeopardized for the same traffic situation if Proposition~\ref{prop:recursive} holds.
	

	\subsection{Safe Crossing}\label{sec:simulation:safe}
\begin{figure}[t]
		\centering
		\includegraphics[width=\columnwidth]{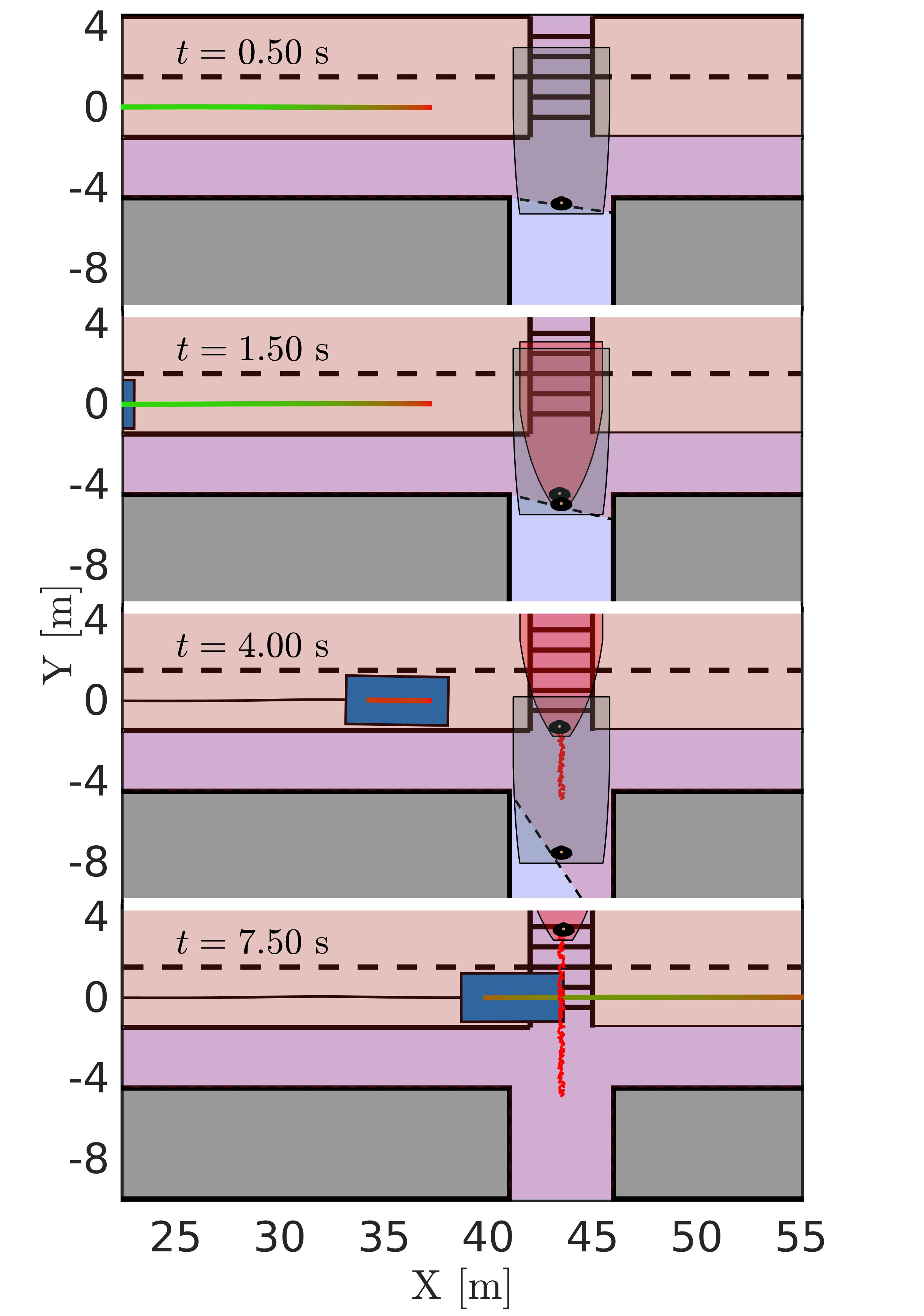}
		\caption{Four different time instances of the simulation environment. The two top panels show that the sensors (shaded region) cannot see behind a wall, however, the vehicle plans a trajectory as if there were a pedestrian behind the corner. The two last panels show that a pedestrian, who previously was not visible, shows up. However, since the vehicle already anticipated that a pedestrian might appear, it manages safely adjust it speed and yield to the pedestrian.}
		\label{fig:no_crash}
	\end{figure}

\begin{figure}[t]
		\centering
		\includegraphics[width=\columnwidth]{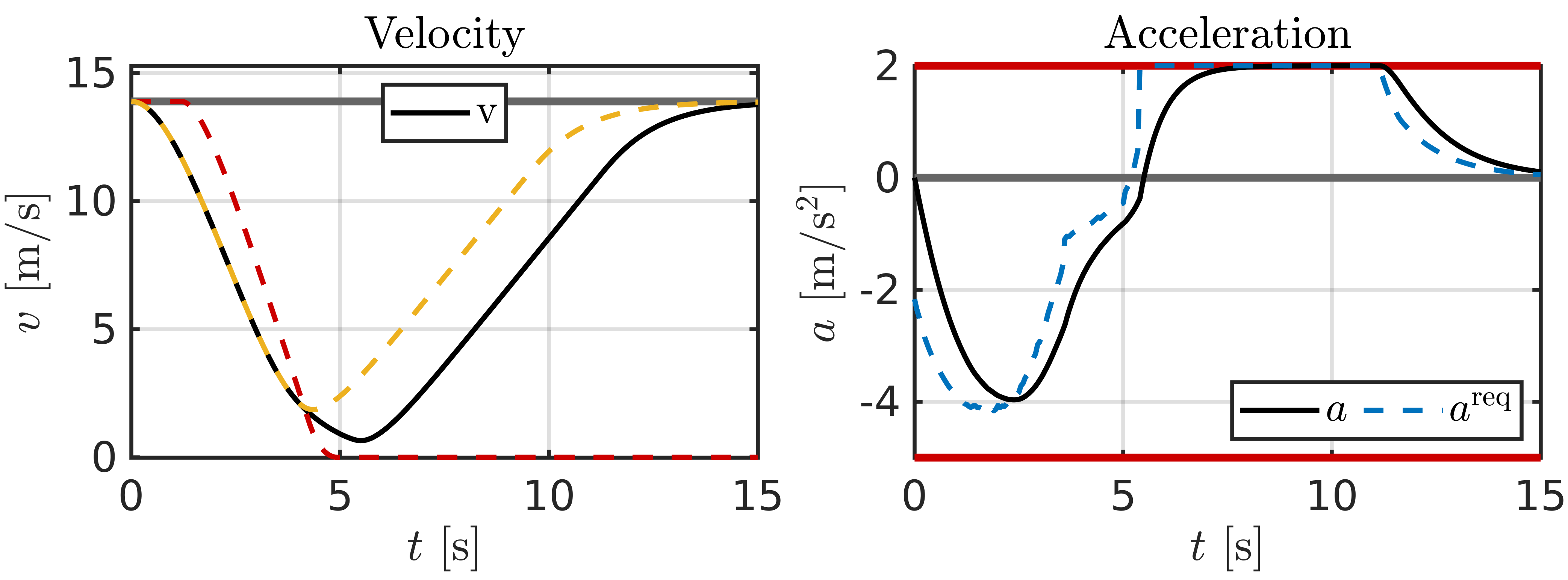}
		\caption{Closed-loop evolution of the safe MPC controller shown in Figure~\ref{fig:no_crash}. The vehicle approaches the intersection by reducing its velocity. After $t=6\ \mathrm{s}$ the pedestrian passes, and the vehicle is free to accelerate again. The yellow dashed line in the left plot shows the closed-loop velocity of the same controller in the case that no pedestrian appears at the intersection, while the red dashed line shows a trajectory comparison of the unsafe controller in Section~\ref{sec:simulation:unsafe}.}
		\label{fig:no_crash_closed}
	\end{figure}

	Figure~\ref{fig:no_crash} shows four time instances of the same MPC controller, however, now occluded pedestrians are anticipated to appear from behind the corner. This is visible from the first two time instances in Figure~\ref{fig:no_crash}, as a ``virtual'' pedestrian is predicted to appear from behind the corner and move towards the intersection. This forces the vehicle to approach the intersection by reducing the velocity, as it can be seen in Figure~\ref{fig:no_crash_closed}. While the vehicle approaches the intersection at a reduced speed (compared to the unsafe controller), the occlusion reduces and the real pedestrian is eventually detected. After $t>6\ \si{s}$ the vehicle is free to pass as it can see that there are no more occlusions, and the real pedestrian has already crossed the road. The yellow dashed line shows the closed-loop velocity of the same MPC controller in the case that no pedestrian appears. Due to the uncertainty, the safe MPC controller, much like a human driver, slows down and is prepared to come to a stop if needed. However, as the vehicle approaches the intersection and the sensor view opens up, it becomes clear that it can cross the intersection safely without stopping unnecessarily.
	
\begin{figure}[t]
		\centering
		\includegraphics[width=\columnwidth]{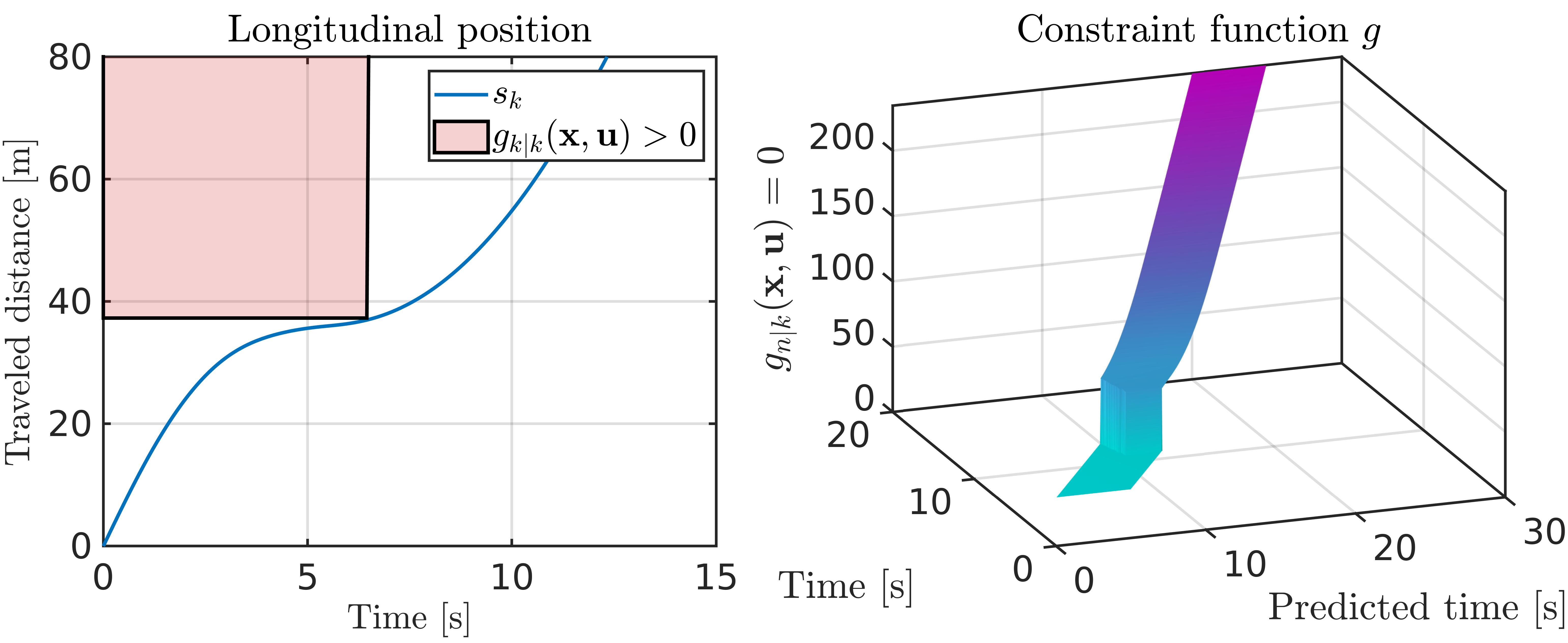}
		\caption{Constraint evolution for the safe MPC controller in Figure~\ref{fig:no_crash}. The left-hand side shows the constraint $g_{k|k}$, while the right-hand side shows the time evolution of the predicted constraint $g_{n|k}$.}
		\label{fig:no_crash_g}
	\end{figure}

	In Figure~\ref{fig:no_crash_g} we can see that the a-priori unknown constraint $g_{n|k}$ satisfies Assumption~\ref{a:unknown} compared to the unsafe controller presented in Section~\ref{sec:simulation:unsafe}. Here it is visible that the constraint does not suddenly shrink, and that $g_{n|k+1}\leq{}g_{n|k}$ for all times.

	This section served as a simple example of the importance that Proposition~\ref{prop:recursive} has on safety. In the next section we deploy the same controller in a real vehicle platform.

	
	\section{Experimental Results}\label{sec:experiments}
	We deployed the safe MPC controller on a full-scale Volvo XC90 T6 petrol-turbo SUV at a closed test track in order to verify the performance in practice. The vehicle offers an actuation interface that accepts longitudinal acceleration requests and steering wheel angle setpoint requests. To access the vehicle CAN bus and sensor system we used the open source software OpenDLV\footnote{See \url{https://opendlv.org/} for more information}, and the middleware library Cluon\footnote{See \url{https://github.com/chrberger/libcluon} for more information}, which enabled reading the sensor data and the sending of actuation requests through the vehicle CAN bus.
	
	\subsection{Platform limitations}
	In order to localize the vehicle we used the onboard Real-Time Kinematic~(RTK) GPS unit and the built-in vehicle IMU sensors to form an estimate of the initial state $\x_k$. However, at the time of testing, no real-time corrections were available, which resulted in a lowered accuracy in the positioning. Furthermore, the interface between OpenDLV and the vehicle showed a time delay of $150\ \si{ms}$ when sending and reading signals related to the steering actuator. Since the steering actuator dynamics were fast, and already accounted for in~\eqref{eq:error_model}, we used dead-reckoning to estimate both the steering angle $\delta$ and steering angle rate $\alpha$. Furthermore, to account for the input delay, we augmented the state space vector in~\eqref{eq:error_model} with additional time-delayed states for the steering angle, similarly to~\cite{batkovic2019real}.
	
	The test vehicle also included a safety feature that limited the steering wheel actuator for different velocities. To capture this limitation, an additional velocity-dependent constraint of the following form was added:
	\begin{equation}
		|\delta_k| \leq{} \Big(10^4\big(1+e^{0.5v}\big)^{-1}+40\Big)\frac{16.8\pi}{180}\si{rad}.
	\end{equation}
	
	The final limitation at the time of testing was that the onboard cameras did not have access to a computer vision software stack, i.e., the detection of the environment, the drivable road, and other road users, was not possible. We therefore  relied only on GPS measurements to localize the vehicle w.r.t. the reference path, and simulated pedestrians that were moving in the environment. While these limitations reduced the experimental scope, we must stress that this setup made it possible to conduct all experiments in a safe and controlled manner. Furthermore, we ought to stress that the main scope of this paper was to deploy the MPC controller~\eqref{eq:nmpc} in a real vehicle platform to test the real-time controller performance, while testing the perception layer of the vehicle is out of the scope of this paper.
	
	\subsection{Controller details}
	Similarly to Section~\ref{sec:simulation}, we consider the parameter values from Table~\ref{tab:param} when formulating~\eqref{eq:nmpc} in the multiple shooting framework with a sampling time of $t_\mathrm{s}=0.05\ \si{s}$, and use a fourth order Runge-Kutta integrator with 5 steps per control interval. We used the stage cost~\eqref{eq:cost_def} and terminal cost~\eqref{eq:terminal_cost_matrix}, together with prediction horizons $N=65$ and $M=100$, and solved the OCP~\eqref{eq:nmpc} using Acados~\cite{Verschueren2019} and HPIPM~\cite{frison2020hpipm} with an SQP-RTI scheme. Note that the only difference between the experimental setup and the simulation setup is the different value on $N$.
	
	To address the mixed-integer problem introduced by constraint $g_{n|k}$, i.e., deciding for which road user to yield and for which to not yield, we formulate~\eqref{eq:nmpc} for each constraint configuration. In the experiments the number of combinations for crossing pedestrians were at most three. To that end, we formulated three OCPs in parallel with each constraint configuration. Then, the control action corresponding to the OCP with the best cost value was applied to the vehicle.
	
\begin{figure}[t]
		\centering
		\includegraphics[width=\columnwidth]{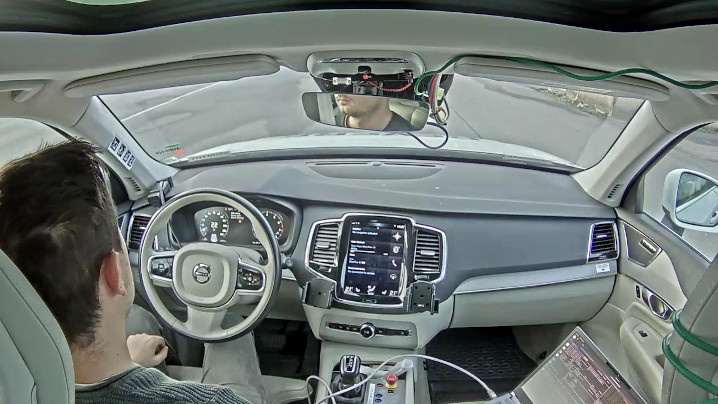}
		\caption{View of the test vehicle that is being controlled by solving Problem~\eqref{eq:nmpc} on a laptop computer, and sending actuation requests over a network connection through OpenDLV.}\label{fig:volvo}
	\end{figure}

	To deploy the controller to the test vehicle, a C++ interface was written to connect Acados and HPIPM with the OpenDLV framework. A Linux laptop~(i9 2.4 GHz, 32 GB RAM) was then used to solve the resulting OCPs in parallel and send the control inputs to the vehicle using OpenDLV, see Figure~\ref{fig:volvo}.

	\subsection{Results}
	In order to evaluate the proposed safe MPC framework, we deploy the controller in a four-way intersection with (simulated) moving pedestrians. In this setting, the vehicle needs to perform a left turn, while safely avoiding any collision with moving pedestrians at two different crosswalks. Indeed, while more general settings would include other road users, we consider only pedestrians in this experiment for simplicity and to be able to clearly illustrate the results.

	\begin{figure}[t]
		\centering
		\includegraphics[width=\columnwidth]{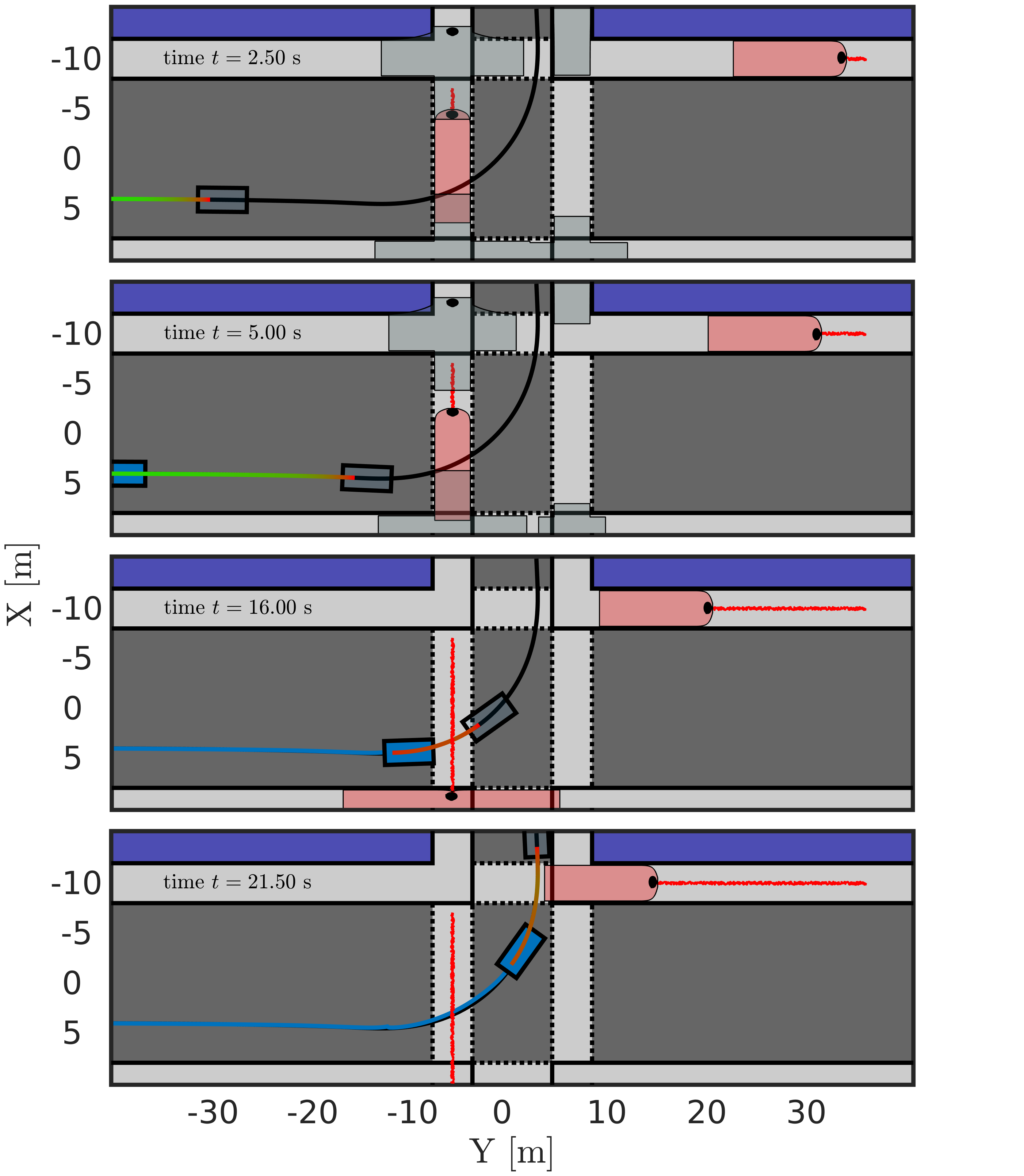}
		\caption{The vehicle bounding box is illustrated by the blue box, while the predicted pedestrian motion is illustrated by the red regions for the real pedestrians and gray regions for the virtual pedestrians. The opaque box represents the terminal state, and the line connecting the vehicle bounding box and the opaque box illustrate the predicted open loop solution. The blue and red lines denote the traveled history of the vehicle and pedestrians, respectively}\label{fig:safe-open}
	\end{figure}
	
	Figure~\ref{fig:safe-open} shows the open-loop solutions across four different time instances. The blue line shows the past positions of the ego vehicle, while the red lines show the past positions of the pedestrians. The red regions denote the future predicted pedestrian uncertainty $\W_{n|k}$ of the measured pedestrians, while the gray regions show the corresponding uncertainty for the ``virtual'' pedestrians. The blue bounding box denotes the ego vehicle at the current time instance, while the opaque box shows the ego vehicle position at time $k+M$, i.e., $\xb[k+M]\in\mathcal{X}_\mathrm{safe}$. The red-green line connecting the blue box and the opaque box denotes the open loop solution in the $xy$-plane, where the red and green colors indicate a low and high velocity, respectively.
	
	The vehicle approaches the first intersection cautiously and yields to the first pedestrian. Then, once the pedestrian has crossed the intersection, the vehicle accelerates up to the next intersection, which also has a pedestrian that moves towards it. In this case, the vehicle approaches the next intersection cautiously and decides to cross only if it can clear the intersection before the pedestrian is predicted to enter. Hence, as the vehicle comes close to the next intersection it realizes that it can clear the intersection safely and therefore accelerates past it in the final frame at $t=21.5\ \si{s}$.
	
\begin{figure}[t]
		\centering
		\includegraphics[width=\columnwidth]{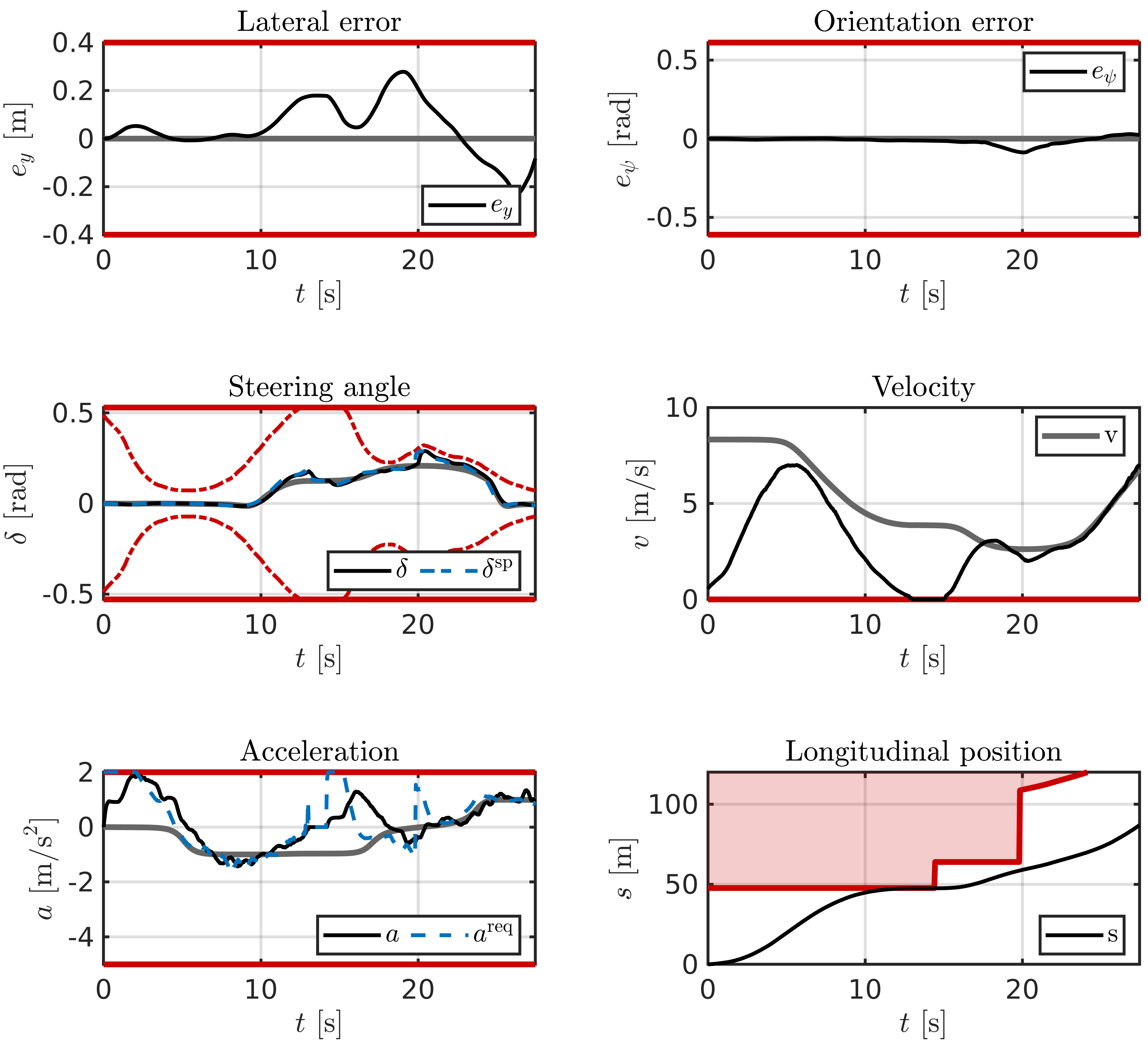}
		\caption{Closed-loop states from Figure~\ref{fig:safe-open}. The black lines represent the corresponding state and input reference, while the blue and orange lines show the state and control inputs.}\label{fig:safe-closed}
	\end{figure}

	Figure~\ref{fig:safe-closed} shows the corresponding closed-loop trajectories for Figure~\ref{fig:safe-open}. Here, it is visible that the vehicle starts far from the reference velocity, which it begins to track. Around time $t=6\ \si{s}$, the vehicle is forced to start deviating from the reference trajectory in order to yield for the first pedestrian and, therefore, comes to a complete stop. After $t=15\ \si{s}$, when the first pedestrian has crossed, the vehicle is allowed to accelerate again and approaches the next intersection, which can be cleared without needing to wait for the other pedestrian to cross. After $t=22\ \si{s}$ it is visible that the velocity is being precisely tracked again.
	
\begin{figure}[t]
		\centering
		\includegraphics[width=\columnwidth]{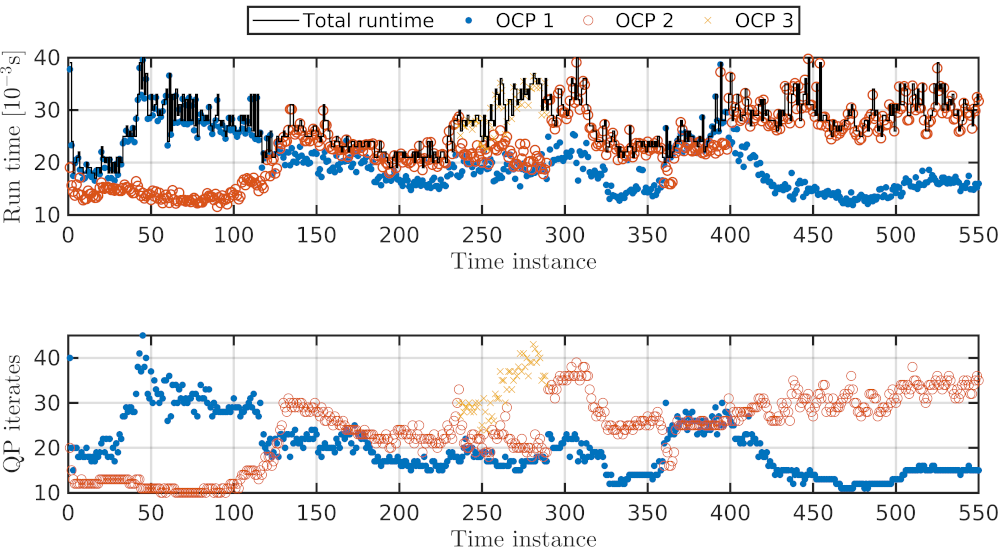}
		\caption{The black line in the top plot shows the total runtime of the framework for each time instance. The markers on the other hand show the solution time for different configurations of~\eqref{eq:nmpc}. The bottom plot shows the number of QP iterations that the solver took for each OCP.}\label{fig:runtime}
	\end{figure}

	Figure~\ref{fig:runtime} shows the solution times of the OCPs and the number of QP iterations it took for the solver for the experiment presented in Figures~\ref{fig:safe-open}-\ref{fig:safe-closed}. Due to the mixed-integer nature of constraint $g_{n|k}$, the number of OCPs that were solved were at most 3 in this particular setting. It is visible that the solution time for all OCPs were at most $40\ \si{ms}$, which was well below the sampling time $t_\mathrm{s}=50\ \mathrm{ms}$.
	
	\section{Conclusions}\label{sec:conclusions}
	In this paper we showed that a safe real-time capable MPC controller can be designed for urban autonomous driving applications. In particular, we have shown how such an MPC framework can be formulated in practice by using a safe set that ensures robust constraint satisfaction at all times, which guarantees recursive feasibility (safety) of the controller. Since one part of the safety guarantees relies on the fact that one has to predict the future motion of other road users, future work will include the investigation and verification of practical and realistic models for the environment prediction, and also consider other road users such as cyclists and other vehicles. Additionally, since the experimental setup in this paper only considered simulated road users, future work will deploy the framework with a sensor suite that enables the online measurements of other road users.
	
	\appendices
	
	\section{Computation of a Stabilizing Terminal Set}\label{appendix:A}
	In this section we show how to derive the terminal set and terminal cost for system~\eqref{eq:error_model} by decoupling the longitudinal and lateral dynamics.

	\subsection{Longitudinal kinematics}\label{sec:appendix_long}
	Since the longitudinal kinematics of~\eqref{eq:error_model} can be written as
	\begin{equation}\label{eq:long_dynamics}
		\matr{c}{\dot{v}\\\dot{a}} = \matr{cc}{0&1\\0&-t_\mathrm{acc}}\matr{c}{v\\a}+\matr{c}{0\\t_\mathrm{acc}}a^\mathrm{req},
	\end{equation}
	and the reference trajectory is assumed to satisfy the system dynamics, i.e., $v^\r$ and $a^\r$ can be generated using~\eqref{eq:long_dynamics}, we can define the following error states and inputs
	\begin{align*}
		e_v := v - v^\r,\ e_a := \dot e_v = a - a^\r, e_{a^\mathrm{req}} := a^\mathrm{req} - a^{\r,\mathrm{req}},
	\end{align*}
	to obtain
	\begin{equation}\label{eq:error_longitudinal_dynamics}
		\matr{c}{\dot{e}_v\\\dot{e}_a} = \matr{cc}{0&1\\0&-t_\mathrm{acc}}\matr{c}{e_v\\e_a}+\matr{c}{0\\t_\mathrm{acc}}e_{a^\mathrm{req}}.
	\end{equation}
	Knowing the state and reference bounds
	\begin{gather}
		0 \leq v \leq \bar{v},\ \underline{a} \leq a \leq \bar{a},\ \underline{a}\leq a^\mathrm{req} \leq \bar{a},\\
		0\leq v^\r\leq 50/3.6,\ |a^\r|\leq 1,\ | a^{\r,\mathrm{req}}|\leq 1.05
	\end{gather}
	we can derive the following error state and input bounds
	\begin{gather}\label{eq:long_error_bounds}
		e_v \leq 5/3.6,\ -4 \leq e_a \leq 1,\ -3.95 \leq e_{a^\mathrm{req}} \leq 0.95.
	\end{gather}

	Then, by applying a zero-order hold discretization of
	~\eqref{eq:long_dynamics} and designing an LQR controller, with cost matrices $Q^\mathrm{LQR}_\mathrm{lon}=\mathrm{diag}({5\times10^{-3}},1)$ and $R^\mathrm{LQR}_\mathrm{lon}=1$, we obtain a feedback gain $K_\mathrm{lon}=[0.0693\ 0.4151]$ that stabilizes~\eqref{eq:error_longitudinal_dynamics}.
	
\begin{figure}[t]
		\centering
		\includegraphics[width=\columnwidth]{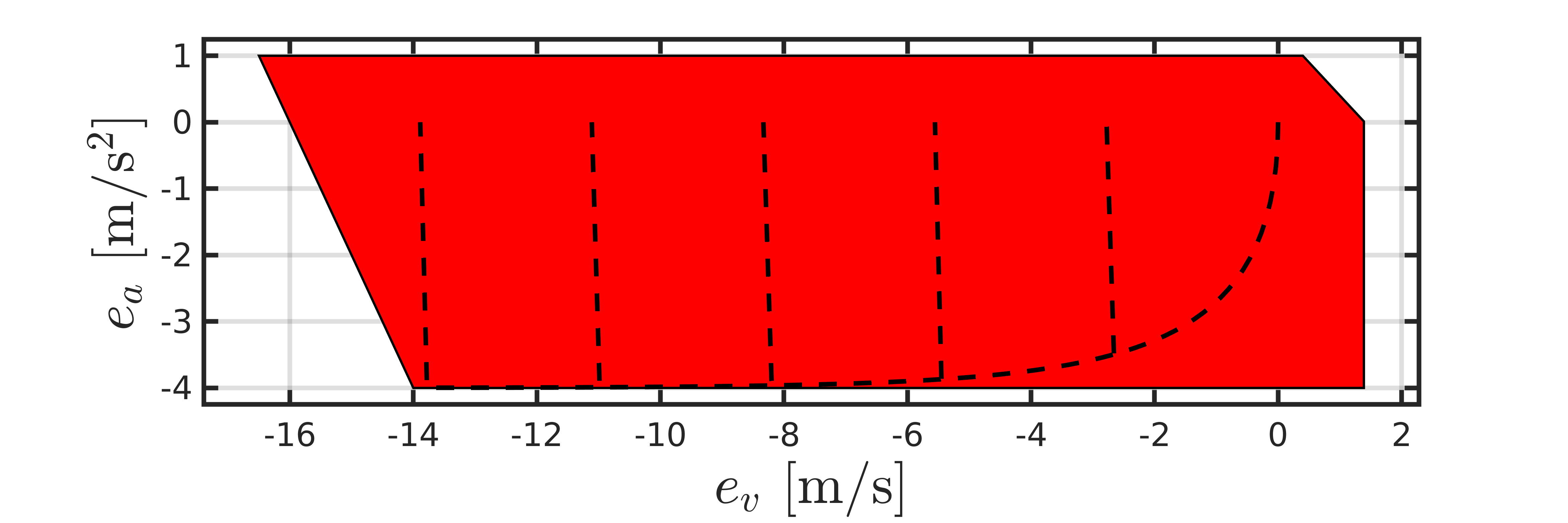}
		\caption{The longitudinal terminal set $\mathcal{X}_\r^\mathrm{lon}$. The black dashed lines show five trajectories where the vehicle starts to brake and applies a constant braking of $a^\mathrm{req}=-4\ \mathrm{m}/\mathrm{s}^2$. Each trajectory starts on the reference, i.e., $v=v^\r$ and $a=a^\r$, where $a^\r=0$ and $v^\r\in\{10,20,30,40,50\}\ \mathrm{km/h}$. }\label{fig:long_terminal_set}
	\end{figure}

	Finally, when computing the invariant set using the geometric approach~\cite{blanchini2008set}, we introduce the additional constraints
	\begin{gather}
		e_v+e_a \leq 1.4,\ -32 \leq 2e_v + e_a, \label{eq:constrict_long_set}
	\end{gather}
	in order to help reduce the complexity. To perform the actual computation, we use the Multi-Parameteric Toolbox~(MPT)~\cite{MPT3} and obtain the following invariant set for the longitudinal kinematics
	\begin{equation}
		\mathcal{X}^\mathrm{lon}_\r(s):=\left \{v,a\ \bigg|\ H_\mathrm{lon}\matr{c}{v-v^\r(s)\\ a-a^\r(s)}\leq b_\mathrm{lon}\right \} ,
	\end{equation}
	where $H_\mathrm{lon}\in\mathbb{R}^{6\times2}$ and $b_\mathrm{lon}\in\mathbb{R}^{6}$. The resulting longitudinal terminal set is shown in Figure~\ref{fig:long_terminal_set}.
	
	\begin{Remark}
		Since~\eqref{eq:long_dynamics} models the longitudinal dynamics of the physical vehicle (the test vehicle in Section~\ref{sec:experiments}) a negative acceleration means that the vehicle applies the brakes to slow down. This implies that the physics of the systems are such that deceleration is only applied if $v>0$, i.e., a braking vehicle will not start reversing when applying brakes and reaching a zero velocity. Formally, this yields a hybrid system. However, due to the simplicity of the hybrid system, this issue can be simply tackled by handling the system as if it were a linear system and by computing the terminal set is as the intersection $\mathcal{X}^\mathrm{lon}_\r(s)\cap \{v,a|v\geq0\}$. To that end, by leaving the lower bound on $e_v$ open, we can construct a low-complexity invariant set, and know that $e_v=v-v^\r\geq- v^\r$ at all times, since the physical vehicle may never reverse. We clarify this further using a set of decelerating trajectories in Figure~\ref{fig:long_terminal_set}.
	\end{Remark}

	As the LQR controller uses a different cost tuning than the MPC controller~\eqref{eq:nmpc}, we compute the associated terminal cost for $\mathcal{X}_\r^\mathrm{lon}$, using YALMIP~\cite{lofberg2004yalmip} and the SDPT3~\cite{toh1999sdpt3} solver to find a $P_\mathrm{lon}$ that satisfies the following Linear Matrix Inequalities~(LMIs)
	\begin{align}\label{eq:long_lmi}
		P_\mathrm{lon} := &\min_P \mathrm{trace}(P)\nonumber\\
		\mathrm{s.t.}\ &P \succ 0,\\
		&(A_\mathrm{lon}-B_\mathrm{lon}K_\mathrm{lon})^\top P(A_\mathrm{lon}-B_\mathrm{lon}K_\mathrm{lon})-P \preceq\nonumber\\
		&\hspace{10em}-(Q_\mathrm{lon}+K_\mathrm{lon}^\top R_\mathrm{lon} K_\mathrm{lon}),\nonumber
	\end{align}
	where $A_\mathrm{lon}$ and $B_\mathrm{lon}$ are obtained through zero-order hold discretization, and $Q_\mathrm{lon}$ and $R_\mathrm{lon}$ correspond to the longitudinal contribution from the cost functions in~\eqref{eq:cost_def}, i.e., $Q_\mathrm{lon} = \mathrm{diag}(1,\ 1)$ and $R_\mathrm{lon}= 4$. Upon solving~\eqref{eq:long_lmi} we obtain
	\begin{equation}
		P_\mathrm{lon}=\matr{cc}{210.78&80.19\\ \phantom{2}80.19&38.29}.
	\end{equation}
	
	\subsection{Lateral kinematics}\label{sec:appendix_lat}
	In this section we derive an invariant set for the lateral kinematics of~\eqref{eq:error_model} for a range of velocities $v\in[1,55/3.6]\ \mathrm{m/s}$, and the following terminal state constraints
	\begin{gather}
		|e_y| \leq \bar e_y = 0.2,\ |e_\psi|\leq \bar e_\psi = 0.1745, |\delta|\leq \bar\delta = 0.5295,\nonumber\\
		|\alpha|\leq \bar\alpha=0.353,\ |\delta^\mathrm{sp}|\leq \bar\delta.
	\end{gather}
	Furthermore, since the reference trajectory is known beforehand, we can easily compute the bounds
	\begin{equation*}
		|\delta^\r|\leq \bar\delta^\rr=0.2079,\ |\dot{\delta}^\r|\leq \bar{\dot\delta}^\r=0.1836,\ |\ddot{\delta}^\r|\leq{} \bar{\ddot\delta}^\r=1.68.
	\end{equation*}
	
	Now, consider the lateral kinematics in~\eqref{eq:error_model} given by
	\begin{equation}\label{eq:lat_error}
		\matr{c}{
			\dot e_y\\ \dot e_\psi\\ \dot \delta\\ \dot\alpha
		}=
		\matr{c}{
			v\sin e_\psi\\
			vl^{-1}(\tan\delta - \tan\beta)\\
			\alpha\\
			w_0^2(\delta^\mathrm{sp}-\delta) - 2w_0w_1\alpha
		},
	\end{equation}
	where we used $\beta := \arctan(\dot{s}/v\tan(\delta^r))$. In order to derive an invariant set for this system, we first need to  account for the nonlinear terms. This is done by introducing the parameters
	\begin{align}
		\varepsilon_\psi := \frac{\sin e_\psi}{e_\psi},&& \varepsilon_\delta := \frac{\tan\delta-\tan\beta}{\delta-\beta},
	\end{align}
	to obtain the following system
	\begin{equation}\label{eq:lat_error_lpv}
		\matr{c}{
			\dot e_y\\ \dot e_\psi\\ \dot \delta\\ \dot\alpha
		}=
		\matr{c}{
			v\varepsilon_\psi e_\psi\\
			vl^{-1}\varepsilon_\delta(\delta - \beta)\\
			\alpha\\
			w_0^2(\delta^\mathrm{sp}-\delta) - 2w_0w_1\alpha
		}.
	\end{equation}
	Then, by denoting the steering error and steering error rate as
	\begin{align}
		e_\delta := \delta - \beta,&& e_\alpha = \dot e_\delta = \dot \delta - \dot \beta = \alpha - \dot \beta,
	\end{align}
	we rewrite the first three lines of~\eqref{eq:lat_error_lpv} as
	\begin{equation}\label{eq:lat_error1}
		\matr{c}{
			\dot e_y\\ \dot e_\psi \\ \dot e_\delta 
		}=
		\matr{c}{
			v\varepsilon_\psi e_\psi\\
			vl^{-1}\varepsilon_\delta e_\delta\\
			e_\alpha
		}.
	\end{equation}
	Finally, in order to deal with the last equation in~\eqref{eq:lat_error_lpv} we change the control input from $\delta^\mathrm{sp}$ to
	\begin{equation}
		\rho := \delta^\mathrm{sp} - \beta - 2w_0^{-1}w_1\dot\beta - w_0^{-2}\ddot\beta,
	\end{equation}
	and define $\nu_\psi := v\varepsilon_\psi$, $\nu_\delta:=vl^{-1}\varepsilon_\delta$, to obtain the Linear Parameter Varying~(LPV) system
	\begin{equation}\label{eq:lat_lpv}
		\dot{\mathbf{e}}=\matr{c}{
			\dot e_y\\ \dot e_\psi\\ \dot e_\delta\\ \dot e_\alpha
		}=
		\matr{c}{
			\nu_\psi e_\psi\\
			\nu_\delta e_\delta\\
			e_\alpha\\
			w_0^2(\rho-e_\delta) - 2w_0w_1 e_\alpha
		},\ \mathbf{e}=\matr{c}{e_y\\e_\psi\\e_\delta\\e_\alpha}
	\end{equation}
	with state $\mathbf{e}$, input $\rho$, and parameters $\nu_\psi,\nu_\delta$, which implicitly depend on $v$, $\varepsilon_\psi$, and $\varepsilon_\delta$. Having derived the LPV system~\eqref{eq:lat_lpv}, our next goal is to construct a polytopic approximation of~\eqref{eq:lat_lpv}, such that the standard methods from the literature can be used to compute an invariant set~\cite{MPT3,blanchini2008set,KVASNICA2015302}. However, in order to do so, we must first understand which possible values the parameters $v$, $\varepsilon_\psi$, and $\varepsilon_\delta$ can take.
	
	To characterize which values $\nu_\psi$ and $\nu_\delta$ can take, we need to first compute the bounds on $\varepsilon_\psi$, $\varepsilon_\delta$. By formulating the worst-case bound on $\beta$ as
	\begin{equation}\label{eq:beta_bound}
		|\beta| = \left |\tan^{-1}\Big(\frac{\tan(\bar \delta^\rr)}{1-\bar e_y \tan(\bar \delta^\rr)l^{-1}}\Big)\right | \leq \bar\beta = 0.2109,
	\end{equation}
	we can compute the following parameter bounds
	\begin{align}\begin{split}\label{eq:varepsilon_bounds}
			0.995\leq \frac{\sin(\bar e_\psi)}{\bar e_\psi} &\leq \varepsilon_\psi \leq \lim_{e_\psi\rightarrow0}\frac{\sin(e_\psi)}{e_\psi} = 1,\\
			1 &\leq \varepsilon_\delta \leq 1.17,
	\end{split}\end{align}
	where the bounds for $\varepsilon_\delta$ were numerically evaluated by gridding $\delta\in[-\bar\delta,\bar\delta]$ and $\beta\in[-\bar\beta,\bar\beta]$. With the bounds from~\eqref{eq:varepsilon_bounds} we construct the following polyhedron
	\begin{align}
		\mathcal{P} := \{v,\nu_\psi, \nu_\delta\ &|\ 1\leq v \leq 55/3.6,\\
		&0.995v\leq\nu_\psi\leq v,\ vl^{-1} \leq \nu_\delta \leq 1.17vl^{-1}\},\nonumber
	\end{align}
	which contains all possible realizations of the parameters in~\eqref{eq:lat_lpv}. Since polyhedron $\mathcal{P}$ now contains 8 vertices and~\eqref{eq:lat_lpv} is expressed using only $\nu_\psi$ and $\nu_\delta$, we realize that we can reduce the number of vertices by projecting $\mathcal{P}$ onto the $\nu_\psi\nu_\delta$-space. In other words we formulate $\mathcal{P}_{\nu_\psi\nu_\delta} = \mathrm{Proj}_{\nu_\psi\nu_\delta}(\mathcal{P})$, and denote by $\mathcal{V}=\{\boldsymbol{\nu}_1,\boldsymbol{\nu}_2,...,\boldsymbol{\nu}_6\}$ the vertices of $\mathcal{P}_{\nu_\psi\nu_\delta}$.
	
	\begin{figure}[t]
		\centering
		\includegraphics[width=\columnwidth]{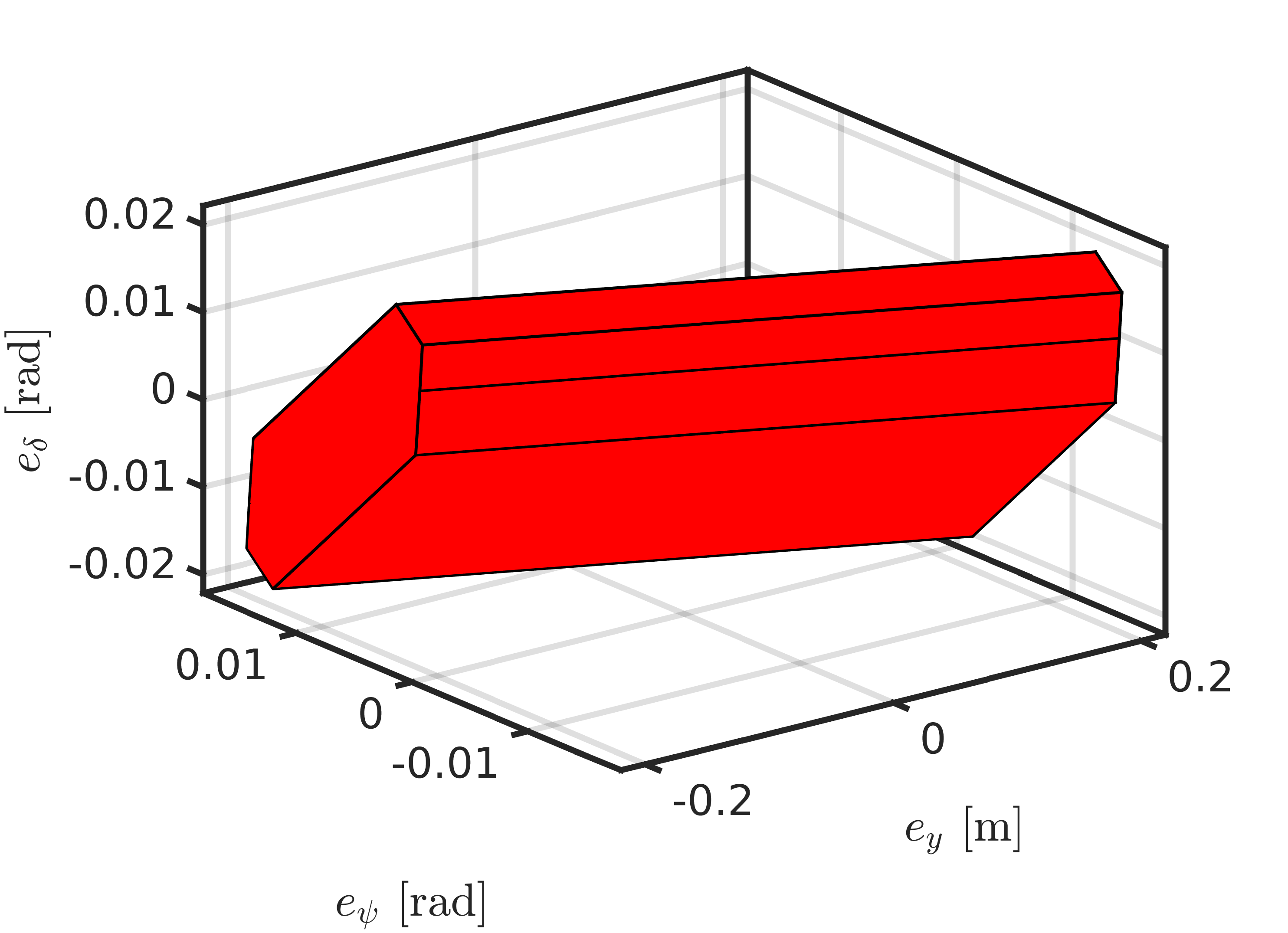}
		\caption{The projection of the lateral terminal set $\tilde{\mathcal{X}}_\r^\mathrm{lat}$ in the $e_y e_ \psi e_\delta$-frame.}\label{fig:lat_terminal}
	\end{figure}
	
	Next, in order to bound the states $e_\delta$, $e_\alpha$ and input $\rho$, we compute the worst-case bounds
	\begin{align}
		|\dot\beta|\leq \bar{\dot\beta} = 0.2013, && |\ddot\beta|\leq \bar{\ddot\beta}=5.9505,
	\end{align}
	numerically by gridding the state space, and compute the following bounds
	\begin{align}\begin{split}
			|e_\delta| &\leq \bar{e}_\delta=\bar\delta - \bar\beta =  0.3186 \\
			|e_\alpha| &\leq \bar{e}_\alpha= \bar\alpha - \bar{\dot\beta} = 0.1517 \\
			|\rho | &\leq \bar \rho = \bar\delta - \bar\beta - 2w_1/w_0\bar{\dot\beta} - 1/w_0^{-2}\bar{\ddot\beta} = 0.2856.
	\end{split}\end{align}

	Knowing the bounds on $(e_y,e_\psi,e_\delta,e_\alpha)$, we now want to construct an invariant set for~\eqref{eq:lat_lpv} that holds for all realizations of the parameters $\boldsymbol\nu\in\mathcal{V}$. We therefore consider the following polytopic approximation~\cite{blanchini2008set} of~\eqref{eq:lat_lpv}
	\begin{equation}
		\Gamma :=  \{ (A,B)\ |\ A = A(\boldsymbol\nu),B = B(\boldsymbol\nu),\ \forall\boldsymbol\nu\in\mathcal{V}\},
	\end{equation}
	where $A(\boldsymbol\nu)$ and $B(\boldsymbol\nu)$ are obtained through zero-order hold discretization. Using the feedback gain $K_\mathrm{lat}$ from an LQR controller for system $(A(\boldsymbol\nu_\mathrm{nom}),B(\boldsymbol\nu_\mathrm{nom}))$ with $\boldsymbol\nu_\mathrm{nom} = [13.89,\ 4.79]^\top$ and tuning $Q_\mathrm{lat}^\mathrm{LQR}=\mathrm{diag}([1,500,1,0.1])$ and $R_\mathrm{lat}^\mathrm{LQR}=10^{-4}$, we can stabilize the polytopic system $\Gamma$ for all $\boldsymbol\nu\in\mathcal{V}$. Then, following the steps in~\cite{KVASNICA2015302} we obtain an invariant set for the lateral dynamics of the form
	\begin{equation}
		\tilde{\mathcal{X}}_\r^\mathrm{lat}(s) = \{ e_y,e_\psi,e_\delta,e_\alpha\ |\ H_\mathrm{lat}[e_y,e_\psi,e_\delta,e_\alpha]^\top\hspace{-0.4em}\leq{}b_\mathrm{lat}\},
	\end{equation}
	where $\tilde{\mathcal{X}}_\r^\mathrm{lat}$ consists of $16$ hyperplanes, i.e., $H_\mathrm{lat}\in\mathbb{R}^{16\times4}$, $b_\mathrm{lat}\in\mathbb{R}^{16}$. The projection of $\tilde{\mathcal{X}}_\r^\mathrm{lat}$ in the $e_y e_\psi e_\delta$-plane is shown in Figure~\ref{fig:lat_terminal}. Since the lateral kinematics are expressed through states $\x_\mathrm{lat}=[e_y,\ e_\psi,\  \delta,\ \alpha]^\top$ for Problem~\eqref{eq:nmpc}, the considered lateral terminal set hence becomes
	\begin{equation*}
		{\mathcal{X}}_\r^\mathrm{lat}(s) = \{ e_y,e_\psi,\delta,\alpha\ |\ H_\mathrm{lat}[e_y,e_\psi,\delta-\delta^\rr,\alpha-\alpha^\rr]^\top\hspace{-0.4em}\leq{}b_\mathrm{lat}\}.
	\end{equation*}

	To find the corresponding terminal cost for $\mathcal{X}_\r^\mathrm{lat}$, we can use Finsler's lemma~\cite{Finsler1936} to derive the following LMIs
	\begin{align}\label{eq:lat_lmi}
		P_\mathrm{lat} := \min_P &\ \mathrm{trace}(P)\nonumber\\
		\mathrm{s.t.}\ &P \succ 0,\\
		&\matr{cc}{P-F& A_\mathrm{lat,cl}(\boldsymbol\nu)^\top P^\top\\ P A_\mathrm{lat,cl}(\boldsymbol\nu) & P}\succ 0, \forall \boldsymbol\nu \in\mathcal{V},\nonumber
	\end{align}
	where $F:=Q_\mathrm{lat}+K_\mathrm{lat}^\top R_\mathrm{lat}K_\mathrm{lat}$, $A_{\mathrm{lat,cl}}(\boldsymbol\nu):=A(\boldsymbol\nu)-B(\boldsymbol\nu)K_\mathrm{lat}$, and $Q_\mathrm{lat}$ and $R_\mathrm{lat}$ correspond to the lateral components of~\eqref{eq:cost_def}, i.e., $Q_\mathrm{lat}=\mathrm{diag}(1,1,10,1)$ and $R_\mathrm{lat}=10$. Note that this corresponds to imposing
	$$ A_\mathrm{lat,cl}(\boldsymbol\nu)^\top P A_\mathrm{lat,cl}(\boldsymbol\nu) - P \preceq -F, $$
	which is quadratic in the parameter $\boldsymbol\nu$. However, this is the Schur complement of the LMI condition in~\eqref{eq:lat_lmi}, which explains the equivalence. Finally, solving~\eqref{eq:lat_lmi} yields
	\begin{align}
		P_\mathrm{lat} = \matr{cccc}{
			325.51 & 593.13 & 97.32 & 1.46\\
			593.13 & 6091.11 & 1979.43 & 29.75\\
			97.32 & 1979.43 & 1159.47 & 17.15\\
			1.46 & 29.75 & 17.15 & 1.28
		}.
	\end{align}
	
	\subsection{Resulting Terminal Set and Terminal cost}
	Following the steps and computing the terminal sets and costs in Appendix~\ref{sec:appendix_long}-\ref{sec:appendix_lat}, we can express the stabilizing terminal set for system~\eqref{eq:error_model} and Problem~\eqref{eq:nmpc} as
	\begin{align}
		\mathcal{X}_\r^\mathrm{s}(s) &:= \{\x\ |\  [v,\ a]^\top\hspace{-0.5em}\in\mathcal{X}_\r^\mathrm{lon}(s), [e_y,\ e_\psi,\ \delta,\ \alpha]^\top\hspace{-0.5em}\in\mathcal{X}_\r^\mathrm{lat}(s)\},\nonumber
	\end{align}
	and the terminal cost as
	\begin{equation}\label{eq:terminal_cost_matrix}
		P=\mathrm{blockdiag}(P_\mathrm{lat},P_\mathrm{lon}).
	\end{equation}

	\bibliographystyle{IEEEtran}
	\bibliography{references}
	\def\dvpic{-1cm}
	\vskip -2\baselineskip plus -1fil
	\begin{IEEEbiography}[{\includegraphics[width=1in,height=1.25in,clip,keepaspectratio]{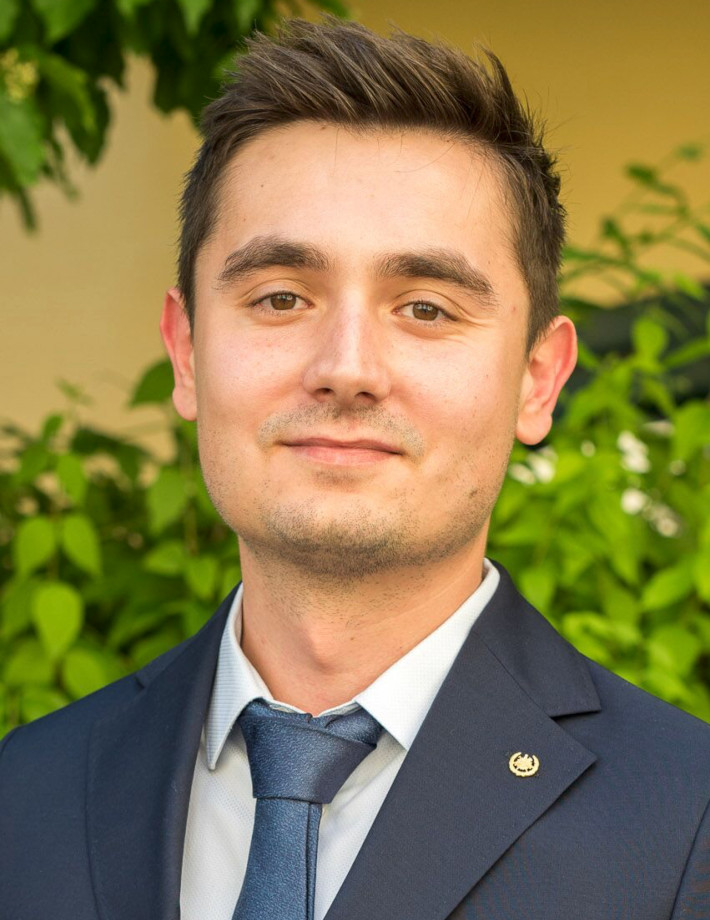}}]{Ivo Batkovic}
		received his M.Sc degree in Engineering Physics in 2016, and his Ph.D. degree in Electrical Engineering in 2022, both from Chalmers University of Technology. He is currently a research engineer at Zenseact in Gothenburg, Sweden, working with decision-making and control for autonomous driving. His research interests include model predictive control for constrained autonomous systems in combination with prediction models for safe decision-making in autonomous driving applications.
	\end{IEEEbiography}
	\vskip -2\baselineskip plus -1fil
	\begin{IEEEbiography}[{\includegraphics[width=1in,height=1.25in,clip,keepaspectratio]{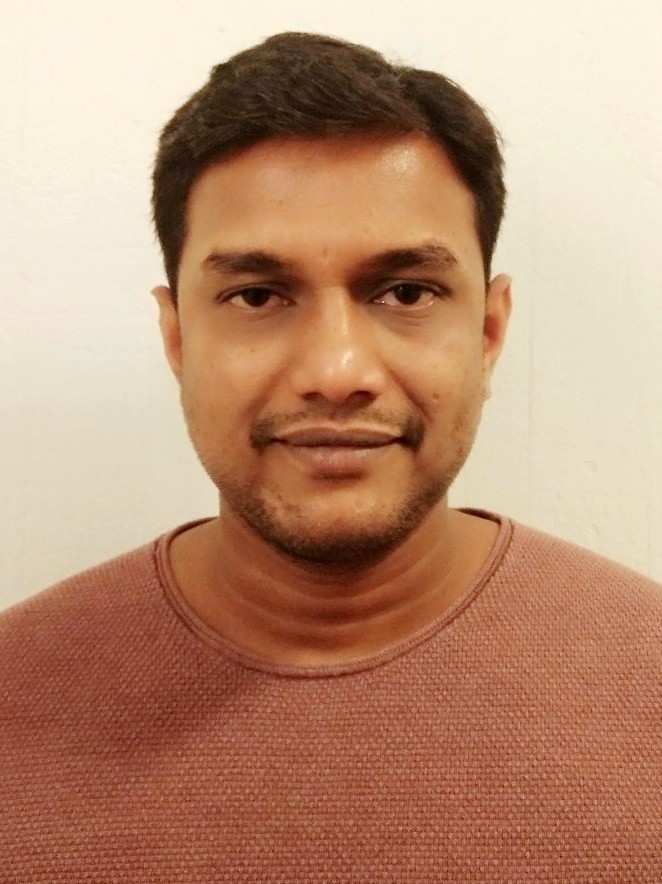}}]{Ankit Gupta} received his Bachelor's degree in Electronics and Telecommunications Engineering in 2010 and his Master's degree in Control Engineering in 2013 from the University of Mumbai, India. He did his doctoral studies at Chalmers University of Technology and is presently working as a control system engineer at Zenseact AB, Sweden. His research interests include invariant sets, reachability analysis, model predictive control, and automotive control.
	\end{IEEEbiography}
	\vskip -2\baselineskip plus -1fil
	\begin{IEEEbiography}[{\includegraphics[width=1in,height=1.25in,clip,keepaspectratio]{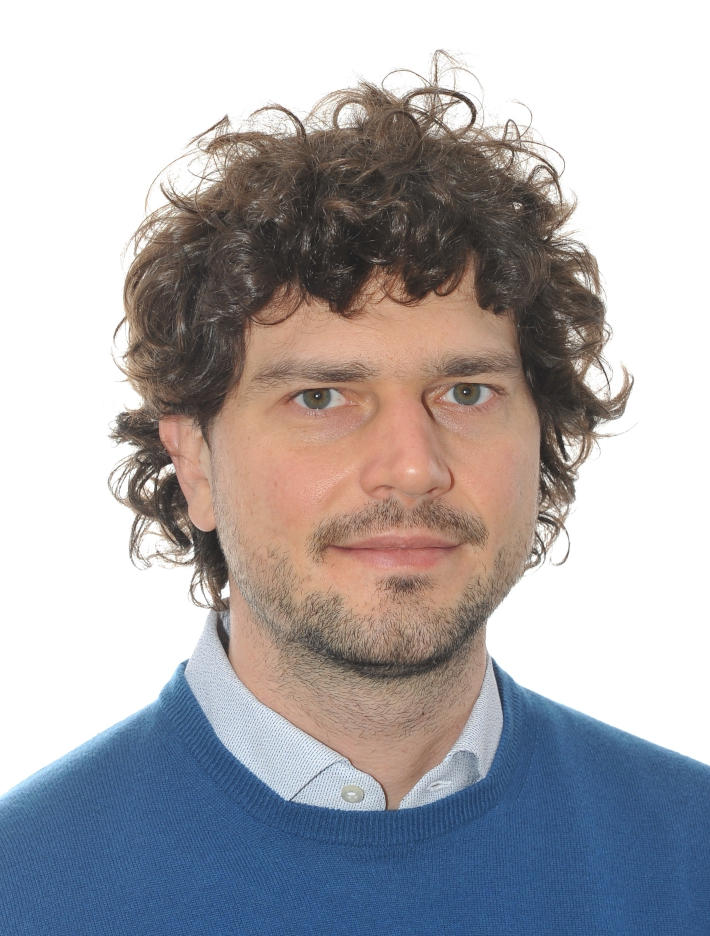}}]{Mario Zanon}
		received the Master's degree in Mechatronics from the University of Trento, and the Dipl\^{o}me d'Ing\'{e}nieur from the Ecole Centrale Paris, in 2010. After research stays at the KU Leuven, University of Bayreuth, Chalmers University, and the University of Freiburg he received the Ph.D. degree in Electrical Engineering from the KU Leuven in November 2015. He held a Post-Doc researcher position at Chalmers University until the end of 2017. In 2018 he became  Assistant Professor at the IMT School for Advanced Studies Lucca where he became Associate Professor in 2021. His research interests include numerical methods for optimization, economic MPC, reinforcement learning, and the optimal control and estimation of nonlinear dynamic systems, in particular for aerospace and automotive applications.
	\end{IEEEbiography}
	\vskip -2\baselineskip plus -1fil
	\begin{IEEEbiography}[{\includegraphics[width=1in,height=1.25in,clip,keepaspectratio]{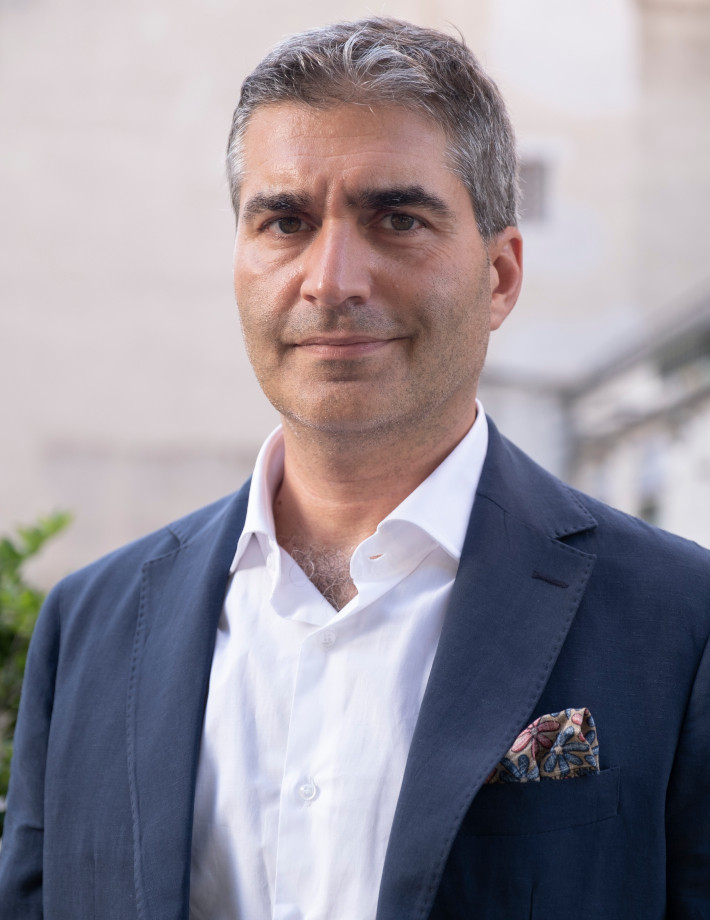}}]{Paolo Falcone}
		received his M.Sc. (“Laurea degree”) in 2003 from the University of Naples “Federico II” and his Ph.D. degree in Information Technology in 2007 from the University of Sannio, in Benevento, Italy. He is Professor at the Department of Electrical Engineering of the Chalmers University of Technology, Sweden. His research focuses on constrained optimal control applied to autonomous and semi-autonomous mobile systems, cooperative driving and intelligent vehicles, in cooperation with the Swedish automotive industry, with a focus on autonomous driving, cooperative driving and vehicle dynamics control.
	\end{IEEEbiography}

\end{document}